\documentclass[11pt]{article}

\usepackage[final]{acl}

\usepackage{times}
\usepackage{latexsym}

\usepackage[T1]{fontenc}

\usepackage[utf8]{inputenc}

\usepackage{microtype}

\usepackage{inconsolata}

\usepackage{graphicx}

\usepackage{amsmath} 
\usepackage{amsfonts}
\usepackage{multirow}
\usepackage{booktabs}
 \usepackage{amssymb}
\usepackage{bbding}
\usepackage{algorithm}
\usepackage{algorithmic}
\title{IntraSlice: Towards High-Performance Structural Pruning with Block-Intra PCA for LLMs}

\author{
 \textbf{Meng Li\textsuperscript{1,2}},
 \textbf{Peisong Wang\textsuperscript{2}},
 \textbf{Yuantian Shao\textsuperscript{1,2}},
 \textbf{Qinghao Hu\textsuperscript{2}},\\
 \textbf{Hongjian Fang\textsuperscript{3}},
 \textbf{Yifan Zhang\textsuperscript{2}},
 \textbf{Zhihui Wei\textsuperscript{1}},
 \textbf{Jian Cheng \textsuperscript{2\rm{*}}}
\\
 \textsuperscript{1}Nanjing University of Science and Technology,
 \textsuperscript{2}Institute of Automation, Chinese Academy of Sciences,\\
 \textsuperscript{3}Beijing National Research Center for Information Science and Technology
\\
 { \small
   \textbf{Correspondence:} 
   \{limeng2024, yuantianshao, gswei\}@njust.edu.cn, 
   \{peisong.wang, yfzhang, jcheng\}@nlpr.ia.ac.cn, 
   }\\
   { \small
   {huqinghao2014@ia.ac.cn}, 
   {nuaalczd@gmail.com}, 
   }
}

\begin{document}
\maketitle
\begin{abstract}
Large Language Models (LLMs) achieve strong performance across diverse tasks but face deployment challenges due to their massive size. Structured pruning offers acceleration benefits but leads to significant performance degradation.
Recent PCA-based pruning methods have alleviated this issue by retaining key activation components, but are only applied between modules in order to fuse the transformation matrix, which introduces extra parameters and severely disrupts activation distributions due to residual connections.
To address these issues, we propose IntraSlice, a framework that applies block-wise module-intra PCA compression pruning. By leveraging the structural characteristics of Transformer modules, we design an approximate PCA method whose transformation matrices can be fully fused into the model without additional parameters.
We also introduce a PCA-based global pruning ratio estimator that further considers the distribution of compressed activations, building on conventional module importance.
We validate our method on Llama2, Llama3, and Phi series across various language benchmarks. Experimental results demonstrate that our approach achieves superior compression performance compared to recent baselines at the same compression ratio or inference speed.

\end{abstract}

\section{Introduction}
Large Language Models have rapidly advanced in recent years \cite{abdin2024phi3, touvron2023llama, zhang2022opt}, demonstrating exceptional performance across a wide range of tasks. However, their enormous parameter scale demands substantial computational resources, which has become a major bottleneck hindering their broader deployment. To address this, numerous model compression techniques \cite{zhu2024survey,xu2023llmsurvey} have been proposed to make LLMs more suitable for resource-constrained environments or edge devices.

\begin{figure}[ht!]
    \centering
    \includegraphics[width=1.0\columnwidth]{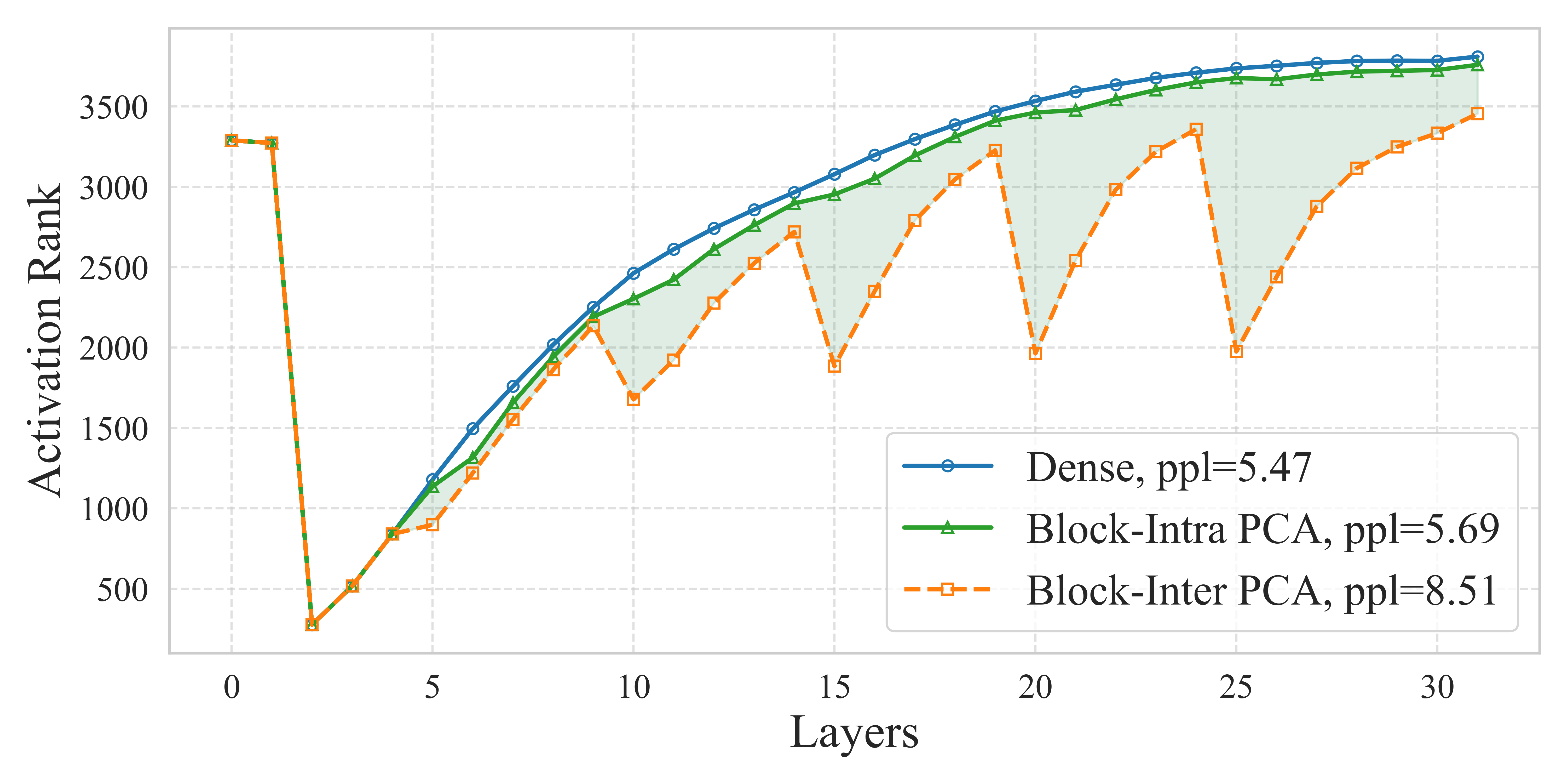} 
    \caption{Comparison of rank changes in layer outputs under PCA compression within modules (Block-intra) vs. between modules (Block-inter) on LLaMA2-7B, applying 50\% sparsity at layers 5, 10, 15, 20, and 25. 
    Lines indicate output ranks, reflecting activation distributions. Block-inter PCA continuously affects the distribution of subsequent activations, keeping them in a low-rank state. In contrast, Block-intra PCA has minimal impact on activation distributions and achieves better results. More detail can be found in Appendix \ref{sec:appendixA}.}
    \label{fig1}
    \vspace{-12pt}
\end{figure}

Structured pruning removes entire components (attention heads or channels) based on model architecture, offering notable acceleration without requiring specialized frameworks or hardware. However, its major drawback is significant performance degradation, limiting practical use. Although recent methods \cite{ashkboos2024slicegpt, gao2024disp} have achieved promising results, performance loss remains a challenge.

Recently, PCA-based methods \cite{ashkboos2024slicegpt} have been proposed to mitigate this issue by applying PCA to inter-module activations and retaining only the most important components during pruning.
This effectively reduces compression error and alleviates pruning-induced performance degradation. However, \textbf{to fuse the transformation matrix into model weight, compression must be applied between modules,} they suffer from two major drawbacks:
(1) The distribution shift in activations by pruning propagates through residual connections and accumulates across layers, as shown in Figure \ref{fig1}, resulting in severe distribution mismatch.
(2) Because activation distributions vary across modules, PCA matrices differ as well, requiring online computation in the residual path and significantly reducing overall acceleration.

To address these limitations, we propose IntraSlice, a PCA-based intra-module pruning method. In transformers-like architectures, \cite{liu2023deja} points out that the output amplitude of individual modules is often much smaller than that of the residual connections. Performing PCA-based compression within modules minimizes its impact on activation distributions and entirely eliminates the need for online computation in the residual connection. However, due to the presence of activation functions and nonlinear components in MHA and FFN, it is difficult to fuse the PCA transformation matrix directly into the weights. 
To resolve this, we combine PCA with the structural characteristics of the model and propose two novel techniques: Adaptive Head Compression with Block-PCA and Progressive Sliced Iterative PCA, which enable effective fusing of the PCA transformation matrix into weights while preserving its strong compression capability. Furthermore, we introduce a global pruning ratio evaluation method that simulates post-transformation activation distributions to allocate distinct pruning ratios across modules in a data-driven manner.

Our main contributions are:
\begin{itemize}
    \item We propose a block-intra PCA solution for transformer modules. This effectively handles the nonlinearity of modules, achieving effective compression and allowing the transformation matrix to be fused into the model weights without introducing additional parameters.
    \item We introduce a global non-uniform pruning ratio evaluation method based on block-PCA, which considers conventional metrics and the distribution of compressed activations to provide more accurate pruning ratio estimation.
    \item We present a novel structured pruning framework for LLMs. Experimental results demonstrate that our method outperforms recent state-of-the-art approaches across a variety of benchmark tasks.
\end{itemize}

\section{Related work}

\subsection{Semi-structured Pruning}
Semi-structured pruning, especially N:M sparsity, is widely used in LLMs \cite{LLM-Barber2024llm, ALPS2024}, enforcing that at least N out of every M weights are pruned to enable efficient matrix operations, enabling efficient execution of matrix-multiply-accumulate operations. SparseGPT \cite{frantar2023sparsegpt} uses Optimal Brain Surgeon (OBS) to prune unimportant weights based on the Hessian and compensates for the induced perturbation. Wanda \cite{wanda2024a} and Plug-and-Play \cite{zhang2024plugandplay} further combine the input and output of weights to judge the importance of weights comprehensively. Prune-Zero \cite{dong2024prunerzero} applies genetic algorithms to evolve importance metrics, treating pruning operations as genetic units. ProxSparse \cite{liu2025proxsparse} adds sparse constraints and obtains the optimal pruning mask through overall optimization. SparseLLM \cite{bai2024sparsellm} and LLM-surgeon \cite{llmsurgeon2024the} extend OBS to global pruning via block-diagonal approximations or inter-block corrections, improving performance but at higher computational cost.

\begin{figure*}[ht]
    \centering
    \includegraphics[width=1.0\textwidth]{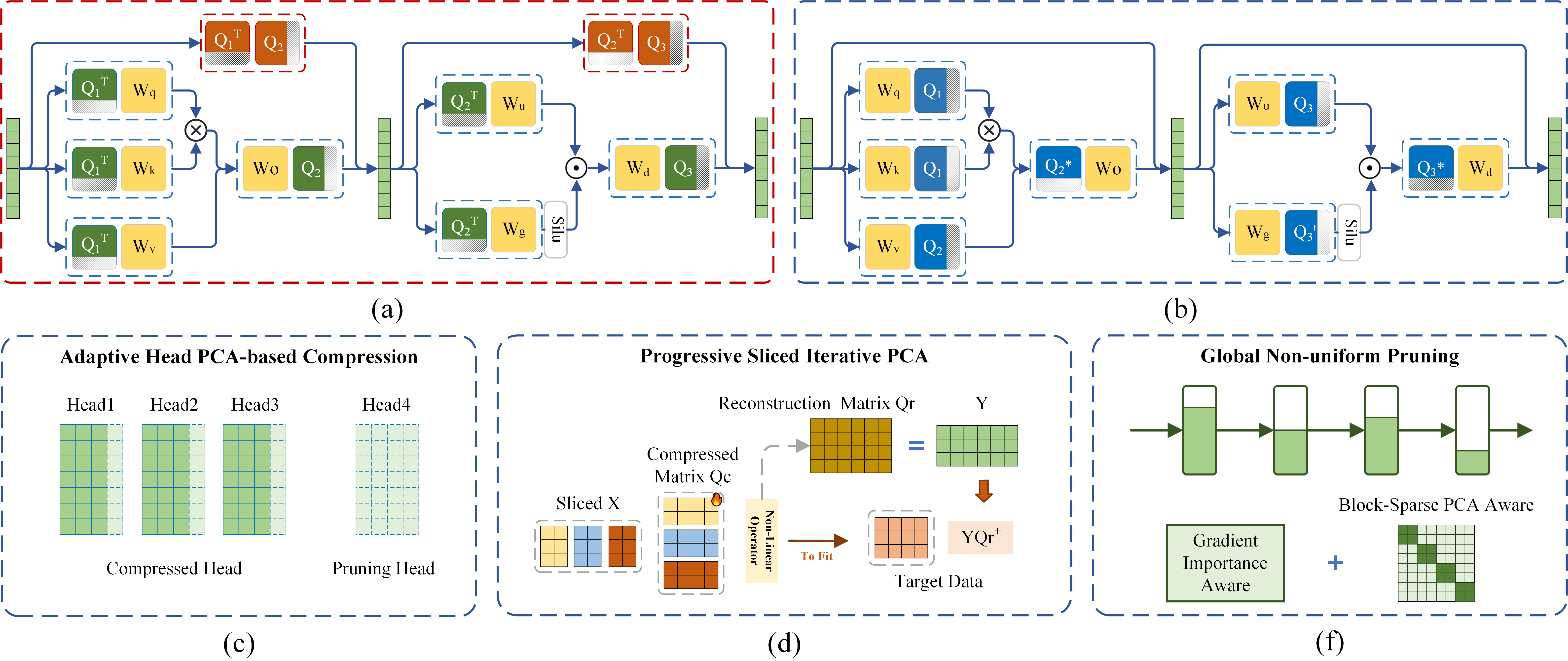} 
    \caption{(a) The existing Inter-PCA pruning frameworks apply PCA compression between modules, which introduces additional computational overhead and error accumulation in residual paths.
(b) Our IntraSlice framework (Intra-PCA) allows full fusion of matrices with less performance degradation.
(c), (d) and (f) are the three components of IntraSlice, respectively.}
    \label{fig2}
    \vspace{-12pt}
\end{figure*}

\subsection{Structured Pruning}
Compared to semi-structured pruning, structured pruning removes entire computational blocks (e.g., attention heads or channels), offering better hardware acceleration \cite{an2024flap, sp3_2024,ashkboos2024slicegpt, gao2024disp}.  FLAP \cite{an2024flap} assigns pruning ratios to different modules via a structured truncation metric. SliceGPT \cite{ashkboos2024slicegpt} uses PCA on inter-module activations, preserving key components via transformation. $SP^{3}$ \cite{sp3_2024} applies PCA within heads and fine-tunes to recover performance. SoBP\cite{wei2024sobp} leverages the gradient of each computational unit’s mask as an importance score, combining with OBS for global optimization. DISP-LLM \cite{gao2024disp} relaxes the constraints imposed by conventional structural pruning methods and calculates the optimal width for the input/output dimensionalitys of each module. Moreover, Coarse-grained approaches like layer \cite{elhoushi2024layerskip, kim2024shortllm} or module pruning \cite{zhang2024finercut, wang2025layer} offer greater speedups but with higher performance loss.

\subsection{Global Non-Uniform Pruning}
Global non-uniform pruning allocates different pruning ratios to modules based on their importance, aiming for optimal sparsity distribution \cite{yin2024outlier, an2024flap}. OWL \cite{yin2024outlier} determines pruning ratios by evaluating each module's ability to reconstruct outlier parameters. BESA \cite{xu2024besa} employs mask learning to enable weight pruning with optimal, module-specific sparsity levels for large language models. SoBP\cite{wei2024sobp} estimates the importance of each computational unit using the magnitude of its mask gradient and performs a global search to assign the most appropriate pruning ratio to each module. Týr-the-Prune \cite{li2025trthepruner} constructs multiple pruning rate candidates for each layer and selects the optimal pruning rate candidate through iterative pruning and optimal search. Although recent methods improve pruning effectiveness, defining reliable evaluation metrics remains challenging.

\section{Method}

For an operation $Y=f(X)$, where $X$ and $Y$ are $ \mathbb{R}^{N \times  D}$. $D$ and $N$ denote data dimensionality and number, respectively. $Q_c$ and $Q_r$ are the corresponding compression matrices, where $Q_c$ is $\mathbb{R} ^{D \times  P}$, and $Q_r$ is $ \mathbb{R}^{P\times D}$. $P$ denotes the target compression dimensionality. The optimization of PCA between $X$ and $Y$ can be defined as:
\begin{equation}
\begin{split} 
\min_{Q_c,Q_r}  \quad  &\left \|  Y - f(X Q_c) Q_r  \right \|_F^2 \\
s.t.&\quad  \left\{\begin{array}{lc}
Q_c^TQ_c=\mathbf{I} \\
Q_rQ_r^T=\mathbf{I}\\
\end{array}\right.
\end{split}
\label{eq:6}
\end{equation}

When $f$ is a linear operator, the problem reduces to a standard PCA formulation. However, the transformer architecture involve complex and diverse nonlinear structures. Specifically, MHA can only fuse block-diagonal matrices, while FFN, due to its stronger nonlinearity, cannot even fuse diagonal matrices. Therefore, we adopt structure-aware strategies tailored to different modules to maximize the reconstruction benefit of PCA. In practice, to improve reconstruction under complex nonlinearities, we removed orthogonality constraints and only limited the compression matrix's amplitude range.

The framework of our method is shown in Figure \ref{fig2}, which contains three main components. (1) \textbf{Adaptive head PCA-based compression}: The compression rate of each head is adaptively adjusted according to the structural characteristics of MHA, while taking into account both speed and accuracy. (2) \textbf{Progressive slicing iterative PCA}: \textit{An optional operation} used to solve the nonlinear PCA optimization problem in FFN. (3) \textbf{Global non-uniform pruning}: Provide different pruning ratios for each block by jointly considering module importance and compressed activation distribution.

\subsection{Adaptive Head PCA-based Compression}
Structured head pruning removes entire attention heads, but this can be suboptimal—pruned heads may still carry useful information, while retained heads may be redundant. Ideally, assigning different compression ratios per head would improve retention of important features, but this breaks attention parallelism and adds time overhead.
To address this, we propose an adaptive structured pruning strategy. IntraSlice removes only completely uninformative heads (pruned heads), while uniformly applying PCA compression to the rest (compressed heads). This balances efficiency and performance without sacrificing parallelism.

\subsubsection{Head reconstruction score.}
Adaptive head compression relies on accurately estimating each head’s importance and its reconstruction ability under different compression ratios. Following Wanda \cite{wanda2024a}, we combine activations and weights to define a channel importance score $I_i$ (Eq. \ref{eq:1}), and obtain head importance $R_h$ by summing its channel scores. $R_h^p$ represents the reconstruction score when head $h$ is pruned and compressed to $p$.
According to PCA theory, the proportion of retained information corresponds to the proportion of the first $p$ largest principal components.
As shown in Eq. \ref{eq:3},  $R_h^p$ is computed based on the eigenvalues $V$ of the Hessian matrix ($X^TX$, $X$ is the output of a head) for head $h$, sorted in descending order. Each eigenvalue reflects the amount of information captured by its corresponding principal component.
\begin{equation}
\vspace{-6pt}
R_h=\sum_{i\in h} I_i\quad; \quad I_i=\left \| X_{:,i} \right \| _F^2\cdot \left \| W_{i,:} \right \| _F^2 \label{eq:1}
\end{equation}
\vspace{-6pt}
\begin{equation}
    R_h^p=R_h\cdot \frac{sum(V_{:p})}{ sum(V)} \label{eq:3}
\end{equation}

\subsubsection{Greedy removal of heads.}
 Following PCA principles, we use reconstruction score maximization as the target. Heads with the lowest scores are greedily removed, and score gains of the remaining heads are evaluated. Let $\kappa$ denote the current set of retained compressed heads, $\left | \kappa  \right |$ be the number of such heads. 
 $P$ denotes the target compression dimensionality of the MHA block, defined as $P=(1-r)D$, where $D$ is the original feature dimensionality and $r$ is the pruning ratio of the MHA block. The $p$ represents the target dimensionality allocated to each remaining compression head, calculated as 
$p=P/ \left | \kappa  \right |$. When a head $h_r$ is pruned, the available compression dimensionality $p$ for the remaining heads increases to $p^*$, as shown in Eq. \ref{eq:5}. The score gain $S_{g}$ is defined as the improvement in the overall reconstruction score resulting from the removal of a head, as shown in Eq. \ref{eq:4}.
If the score gain $S_{g}$ after removing a candidate head $h_r$ is greater than zero, it is considered beneficial to remove $h_r$. Otherwise, $h_r$ is retained as a compressed head.

This greedy iterative process efficiently approximates the optimal reconstruction score. Heads with negligible contribution are pruned, while the rest are compressed via PCA to retain key components. This avoids the substantial information loss caused by blindly removing entire heads.
\begin{equation}
\vspace{-6pt}
    S_{g}=\sum_{h\in \kappa,h\ne h_r}R_h^{p^*}-\sum_{h\in \kappa } R_h^p \label{eq:4} 
\end{equation}
\begin{equation}
p^*=p+\frac{p}{\left | \kappa \right |-1 }=\frac{P}{\left | \kappa \right |-1 }  \label{eq:5}
\vspace{-6pt}
\end{equation}

\begin{figure}[!t]
    \centering
    \includegraphics[width=0.95\columnwidth]{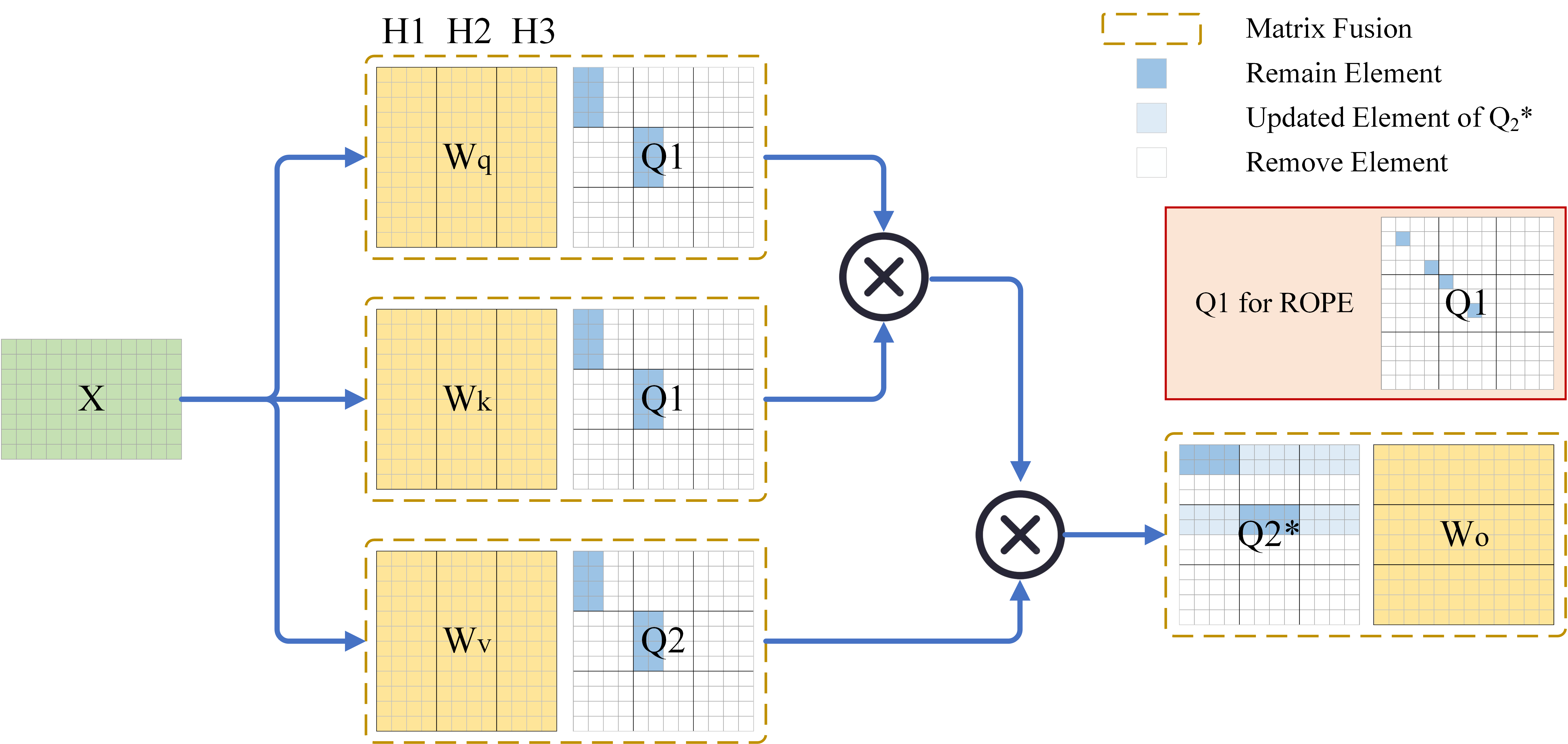} 
    \caption{Schematic diagram of adaptive head PCA-based compression structure pruning and weight fusion.}
    \label{fig3}
\end{figure}
\subsubsection{Structured pruning and weight fusion.}
The attention module involves two PCA compression points: between value and output projections, and between query and key for attention score computation.
However, the attention structure is inherently nonlinear, which complicates the integration of PCA transformation matrices.
To address this, each head is treated as an independent compression unit as illustrated in Figure \ref{fig3}. The PCA transformation matrix $Q_1$ calculated by query and key is fused directly into the $W_q$ and $W_k$, respectively. Specifically, to be fused into the value projection weight $W_v$, $Q_2$ is required to be a block-wise matrix. 
Importantly, it allows the transformation matrix $Q_2^T$ fused into the $W_o$ to be fully dense, meaning that the heads can compensate for others, as in Eq. \ref{eq:20}, where $X$ is the input of $W_o$. 
\begin{equation}
\vspace{-6pt}
Q_2^* = ((XQ_2)^T(XQ_2)+\lambda\mathbf{I})^{-1}(XQ_2)^TX  \label{eq:20}
\end{equation}
In models with Rotary Position Embedding (RoPE), the added nonlinearity complicates attention score computation. To enable efficient fusion, $Q_1$ is simplified into a pairwise channel selection matrix. Experiments show this has little impact on compression performance. More details of matrix fusion can be found in Appendix \ref{sec:appendix_rope}.

\subsection{Progressive Sliced Iterative PCA}
Due to nonlinear activations and complex FFN structures, compression matrices in MLPs can't be directly fused into model weights. For such nonlinear compression problems, traditional iterative PCA requires loading full activation data repeatedly \cite{pca2011principal}, which is costly—e.g., in LLaMA2-7B, extracting enough components may need thousands of iterations and heavy CPU-GPU transfers. Furthermore, constructing principal components using the entire activation dimensionality often fails to yield effective results due to the high complexity of the data.
To address this issue, we propose a PCA method based on data slicing and progressive iteration (denoted as \textbf{IterPCA}). Unlike traditional PCA that splits along the compression dimension, our method slices by data dimension and incrementally updates the transformation matrix, reducing full data loading and lowering access and computation costs.

\subsubsection {Iterative optimization with data slices.} 
As shown in Figure \ref{fig2} (d), to further accelerate the process, we reformulate the optimization objective: instead of optimizing the reconstruction matrix $Q_r$, let $Y^r=YQ_r^{+}$ be the target, where $Q_r^+$ is the pseudo inverse of $Q_r$, $Y^r$ is $ \mathbb{R}^{N \times  P}$. Here, $Q_r$ can be obtained via $Y$ as a suboptimal reconstruction matrix.
For the optimization of $Q_c$, we draw inspiration from parallel acceleration in matrix computation, where a large dot-product operation can be decomposed into a sum of dot-products over smaller data blocks.
We divide $X$ into several $d$-dimensional parts, where $d$ is the slice dimensionality of $D$ after partitioning. In the $k^{th}$ data slice, the corresponding optimization of the $k^{th}$ part of $Q_c$ is :
\begin{equation}
\small
\begin{split} 
\begin{split} 
\min_{Q_{c}}  \quad & \left \|  Y^r  - f(X_{[:,(k-1)d:kd]} Q_{c[(k-1)d: kd]}+C)\ \right \|_F^2 \\
&\quad \quad C=\sum_{i=1}^{k-1} {X_{[:,(i-1)d:id]} Q_{c[(i-1)d: id]}} 
\end{split}\\
\end{split}
\label{eq:7}
\end{equation}
\vspace{-6pt}
\begin{equation}
\small
\begin{split} 
Q_r = (f(XQ_c)^Tf(XQ_c)+\lambda\mathrm{I})^{-1}f(XQ_c)^TY
\end{split}
\label{eq:8}
\end{equation}
Where $C$ is the cumulative sum of the previous $k-1$ steps and has the same size as $Y^r$.
Once $Q_c$ is optimized, we recompute $Q_r$ with Eq. \ref{eq:8}.
To achieve better results, we iteratively apply Eq. \ref{eq:7} and Eq. \ref{eq:8}, updating $Q_c$ and $Q_r$ in sequence. $Q_c$ is initialized as a channel selection matrix based on amplitude, while $Q_r$ is initialized from $Q_c$ using Eq. \ref{eq:8}. 
In practice, iterative process is \textit{an optional step} applied only when the current layer has a relatively high pruning ratio (around 10\% of the layers). Mostly, we directly use the initialization results of $Q_c$ and $Q_r$, so its overall time cost remains low.

\begin{table*}[ht]
    \centering
    \small 
    \setlength{\tabcolsep}{2mm}
    \renewcommand{\arraystretch}{0.85}
    {
    \begin{tabular}{c|c|cc|cc|cc|cc|cc}
         \toprule
         \multirow{2}{*}{\textbf{Sparsity}} & \multirow{2}{*}{\textbf{Method}} & \multicolumn{2}{|c}{\textbf{Llama2-7B}} &  \multicolumn{2}{|c}{\textbf{Llama2-13B}} &  \multicolumn{2}{|c}{\textbf{Llama2-70B}} &   \multicolumn{2}{|c}{\textbf{Llama3-8B}} & \multicolumn{2}{|c}{\textbf{Phi-3-Medium-4k}} \\
            
            \cline{3-12}
            & & PPL$\downarrow$  & Avg$\uparrow $  & PPL$\downarrow $  & Avg$\uparrow $  & PPL$\downarrow $  & Avg$\uparrow $  &PPL$\downarrow $  & Avg$\uparrow $ & PPL$\downarrow$  & Avg$\uparrow $    \\
         \midrule
         
         0\% & Dense &  5.47& 66.69 & 4.88 &69.24 & 3.32 &73.61 &6.13  &70.00 & 4.29 & 74.75\\
         \midrule
         \multirow{6}{*}{20\%} & SliceGPT   &  6.84 & 54.25 &  6.06 & 56.78 &  4.46 & 69.60 & 10.93 &48.10 & 6.19 & 66.43 \\
                               & Wanda   &  7.38 & 61.25 &  6.66 & 61.00 &  4.1 & 70.86 & 122.41 &37.53 & 6.82 & 69.52 \\
                               & FLAP       &  7.16 &  56.62 & 6.31 & 61.55 & 4.12  & 71.60 &  -- & -- & -- & --\\
                               & SoBP      &  6.53 &  63.27 & 5.62 & 67.73 & 3.88  & 71.24 &    8.74 &  63.56  & 6.27& 72.29\\
                               & SVD-LLM  & 8.52  & 49.60 & 6.78  & 58.56 & --    & -- & 47.00 & 45.43 & 7.16  & 68.02 \\
                               & IntraSlice &  \textbf{6.27} & \textbf{63.73} & \textbf{5.48}  &  \textbf{67.94}& \textbf{3.85} & \textbf{72.95} & \textbf{8.27} & \textbf{65.21}& \textbf{5.80} & \textbf{73.06}\\
         \midrule
            \multirow{6}{*}{30\%} & SliceGPT   & 8.64 &46.70 & 7.44& 50.10 & 5.41 & 61.61 & 17.02 & 41.40 & 7.52 & 56.82 \\    
                                & Wanda   & 9.17  & 56.28 & 10.14 & 44.72 & 4.77 & 69.90 & 271.71 & 36.50  & 10.00 & 62.41 \\
                               & FLAP       &   8.85  &   50.91 & 7.57 & 57.27 & 4.82  & 69.68 & -- & -- & -- &--\\
                               & SoBP      &  7.58&  59.15&  6.27&  66.82 & 4.36&70.30 &      10.32    &  58.60  & 7.05 & 67.52 \\
                               & SVD-LLM  & 10.95 & 45.14 & 8.21  & 52.16 & --    & -- & 101.56 & 40.80 & 8.22  & 61.62 \\
                               & IntraSlice &  \textbf{7.11} &  \textbf{60.49} &  \textbf{5.96} &  \textbf{67.13} &\textbf{4.34}  & \textbf{72.27} &\textbf{10.25} & \textbf{60.65} &\textbf{6.71} &\textbf{69.73} \\
        \midrule
        \multirow{6}{*}{40\%} & SliceGPT   &  12.80 &41.47 &10.60 &44.31 & 7.08 &52.00 & 30.80 & 37.39 & 10.19 & 45.25  \\
                               & Wanda   & 14.33 & 43.05  & 21.34 & 41.35  & 5.82  & 66.36  & 4258.41 & 34.68  & 20.68 & 53.52 \\
                               & FLAP       &  11.49 &  48.70 &  9.07&  53.18 &  6.24&  67.96 &   -- & --  & -- & --\\
                               & SoBP      &  9.28 & 56.06 & 7.39 & 60.86 &  4.96 & 68.58 & 12.48 &  \textbf{52.71}  & 8.02 & 61.15 \\
                               & SVD-LLM  & 16.58 & 39.37 & 11.26 & 44.61 & --    & -- & 207.99 & 36.71 & 10.76 & 52.51 \\
                               & IntraSlice &  \textbf{8.39} & \textbf{56.38} &\textbf{6.92}  &  \textbf{62.24}& \textbf{4.91} & \textbf{69.99} & \textbf{12.28} & 51.10&\textbf{7.90} &\textbf{64.90} \\
         
         \bottomrule
    \end{tabular}}
    \caption{Comparison of model compression results of different methods. The bold ones indicate the best results. PPL is the result on wikitext2, and Avg is the average result of 7 zero-shot tasks.}
    \label{table1}
\end{table*}

\subsection{Global Non-uniform Pruning}
Global non-uniform pruning aims to assign different pruning ratios to blocks by evaluating the importance of computational units across layers. In SoBP\cite{wei2024sobp}, a mask is applied to each unit, and its gradient with respect to the final loss is used as an importance metric.
To enhance accuracy, modern strategies often incorporate data compensation techniques such as OBS \cite{hassibi1993obs}, which can alter the activation distribution used during pruning-rate estimation. Therefore, it is essential to consider compensation effects when estimating pruning ratios.
To this end, we propose a PCA-based global non-uniform pruning-rate estimation method that accounts for both unit importance and the impact of PCA transformations.

\subsubsection{Mask-based importance evaluation.}
Following the importance calculation strategy introduced in SoBP\cite{wei2024sobp}, we apply a mask to each computational unit.  The mask application scheme in MHA and FFN modules is illustrated in Eq. \ref{eq:10}. $X_h^{l,i}$ is the output of the $i^{th}$ head in $l$ layer. $Cat$ is the concatenation operation, $M_h^l$ is the mask of MHA for $l$ layer. Instead of setting one mask for the entire head, our method enables more fine-grained pruning within MHA module, setting a mask for each channel, $M_h^l$ is $ \mathbb{R}^{D}$, $D$ is the hidden size. The mask of the FFN module is set similarly, $M_f^l$ is $ \mathbb{R}^{D_{inter}}$, $D_{inter}$ is the intermediate size.
\begin{equation}
\small
\begin{split}
      X_{mha}^l = (Cat(X_h^{l,1},X_h^{l,2},...,X_h^{l,H}) \circ M_{h}^l)W_o^l \\
    X_{ffn}^l = ((X_{mha}^l W_u^l)\circ \sigma (X_{mha}^lW_g^l)) \circ M_{f}^l)W_d^l 
\end{split}
  \label{eq:10}
\end{equation}
According to the Taylor expansion, for layer $l$, the change in the final loss $\mathcal{L}$ for the mask $m_i^l$ can be approximated as $g_i^l(m_i^l-1)$, where $g_i^l$ is the gradient of $m_i^l$ when the mask is 1, $m_i^l$ is the $i^{th}$ mask of $M_h^l$ or $M_f^l$. Then the importance $I_i^l$ of this computational unit can be approximated with Eq. \ref{eq:11}, where $\mathcal{D}$ is calibration data.
\begin{equation}
\begin{split} 
    I_i^l = (\mathcal{L} (m_i^l,&\mathcal{D})-\mathcal{L} (1 ,\mathcal{D}))^2 \approx (g_i^l)^2 \\
    & m_i^l = 0
\end{split} 
    \label{eq:11}
\end{equation}

\subsubsection{Sparse PCA-aware importance correction.}
Since our method uses approximate PCA, we adopt sparse PCA to better approximate the transformation and accelerate computation. For layer $l$, let $Q_s^l$ denote the transformation matrix derived from the block-PCA of activations $X^l$. $Q_s^l$ is a block diagonal matrix ($\mathcal{R}^{D\times D}$ for MHA and $\mathcal{R}^{D_{inter}\times D_{inter}}$ for FFN). After transformation, the activation becomes $X^lQ_s^l$, and the corresponding weight becomes ${Q_s^l}^TW^l$. According to the backpropagation chain rule, the gradient of the mask $M_h^l$ or $M_f^l$ is transformed as $g^lQ_s^l$, where $g^l$ represents the original gradient of the corresponding mask of the $l$ layer before PCA transformation. The corrected importance score of single mask $m_i^l$ is then calculated based on this adjustment with Eq. \ref{eq:12}.
\begin{equation}
    I_i^l = ((g^lQ_s^l)_i)^2 \label{eq:12}
\end{equation}

Due to structural differences, the importance of MHA and FFN is not directly comparable. We generally assign the same pruning rate to both, and search for the optimal pruning rate for MHA or FFN respectively, by maximizing the sum of the importance scores $I$ of the retained computing units. To control pruning tendency, we introduce a bias $\lambda_b$, scaling the MLP pruning ratio to $\lambda_b \times prune\_ratio$, while adjusting the MHA pruning ratio to maintain overall sparsity. This offers a more intuitive balance than tuning per-component importance \cite{wei2024sobp}. In practice, setting $\lambda_b \approx 1$ is often sufficient (see Appendix \ref{sec:appendix_detail} for details).

\section{Experiments}

\subsection{Experiment Setup}
\subsubsection{Models.} 
We evaluate our IntraSlice on several LLMs with a transformer architecture. Specifically, we consider models from the LLaMA-series \cite{touvron2023llama2, dubey2024llama3}: LLaMA2-7B, LLaMA2-13B, LLaMA2-70B, and LLaMA3-8B; Phi-series \cite{abdin2024phi3, javaheripi2023phi2}: Phi-3-Medium-4K.

\subsubsection{Baselines \& tasks.} 
We compare our method with several recent state-of-the-art structured pruning approaches, including Wanda \cite{wanda2024a}, SliceGPT \cite{ashkboos2024slicegpt}, FLAP \cite{an2024flap}, SVD-LLM \cite{wang2025svdllm}and SoBP \cite{wei2024sobp}. 
In addition, we also compared with some more expensive methods( consume more data or longer time for pruning), such as DISP-LLM \cite{gao2024disp}, t\'yr-the-Pruner \cite{li2025trthepruner} and LLM-surgeon \cite{llmsurgeon2024the}. 
Following SoBP\cite{wei2024sobp}, we adopt the llm-eval-harness \cite{gao2021llm-eval} to evaluate our method and other compared methods on 7 widely-used zero-shot tasks: ARC-c, ARC-e \cite{clark2018data-arc_e_c}, WinoGrande \cite{sakaguchi2020data-winogrande}, BoolQ \cite{clark2019boolq}, HellaSwag \cite{zellers2019data-hellaswag}, OpenBookQA \cite{openbookqa-2018-suit} and PIQA \cite{bisk2020data-piqa}.

\begin{table*}[!t]
\centering
\small
\setlength{\tabcolsep}{4pt}
\renewcommand{\arraystretch}{0.85}
\begin{tabular}{c|c|c|ccccccc|c}
\toprule
\textbf{Sparsity} & \textbf{Model} & \textbf{PPL} & \textbf{WinoGrande} & \textbf{PIQA} & \textbf{OBQA} & \textbf{HellaSwag} & \textbf{BoolQ} & \textbf{ARC\_e} & \textbf{ARC\_c} & \textbf{Avg} \\
\midrule
 0\% & Llama2-13b & 4.88 & 72.3 & 80.52 & 45.2 & 79.38 & 80.58 & 77.53 & 49.23 & 69.25 \\
\midrule
\multirow{2}{*}{50\%}  & SoBP & 9.22  & \textbf{64.48} & 53.92 & 35.40 & \textbf{58.13} & 69.66 & 28.41 & 25.43 & 47.92 \\
  & IntraSlice &  \textbf{8.75}  & 63.30 & \textbf{66.87} & \textbf{37.00} & 56.78 & \textbf{69.85} & \textbf{58.38} & \textbf{33.53} &  \textbf{55.10} \\
\midrule
\multirow{2}{*}{60\%} 
  & SoBP       & 15.62           & \textbf{59.27} & 56.53        & 28.40             & 45.60      & \textbf{66.27}  & 33.04 & 26.96 & 45.15 \\
  & IntraSlice &  \textbf{11.36} & 57.70         & \textbf{62.30} & \textbf{32.60 }& \textbf{46.40} & 61.71        & \textbf{48.06} & \textbf{29.35} &  \textbf{48.30}  \\
\bottomrule
\end{tabular}
\caption{Comparison of model compression results with high sparsity.}
\label{table_high_sparse}
\end{table*}

\begin{table}[!b]
    \centering
    \small
    \setlength{\tabcolsep}{2mm}
    \renewcommand{\arraystretch}{0.8}
    {
    \begin{tabular}{c|c|c|c|c}
         \toprule
         \textbf{Sparsity} &\textbf{Method} & {\textbf{E-T}}  &  \textbf{PPL} &   \textbf{Avg} \\   
            
         \midrule
         
           0\%        & Llama2-7B& --& 5.12 &     68.99\\
         \midrule    
         \multirow{3}{*}{30\%} & LLM-Surgeon&  \XSolidBrush &7.83 &  59.03   \\
                               & DISP-LLM   &  \Checkmark  &6.85 &   58.10    \\
                               & týr-the-Pruner &  \XSolidBrush &7.00 & 57.92 \\
                               & IntraSlice & \XSolidBrush &\textbf{6.61}  &   \textbf{62.13}   \\
         \midrule
         \multirow{3}{*}{50\%} & LLM-Surgeon& \XSolidBrush &15.38 &   45.68    \\
                               & DISP-LLM   & \Checkmark &\textbf{9.84}   & 46.72    \\
                                & týr-the-Pruner &  \XSolidBrush &10.44 & 49.10 \\
                               & IntraSlice & \XSolidBrush &10.11 &  \textbf{50.85}  \\
       \toprule

        0\% & Llama2-13B& -- &4.25	&	71.79 \\
         \midrule

         \multirow{4}{*}{30\%} & LLM-Surgeon  & \XSolidBrush&6.21 & 65.34    \\
                                & DISP-LLM  &\Checkmark  &5.77 & 63.07    \\
                                & týr-the-Pruner &  \XSolidBrush &6.05 & 64.75 \\
                               & IntraSlice & \XSolidBrush &\textbf{5.63} &	\textbf{68.27}  \\ 
                               \midrule

        \multirow{4}{*}{50\%} &  LLM-Surgeon  & \XSolidBrush &9.43 &	55.24    \\
                                & DISP-LLM & \Checkmark  & \textbf{7.11} &54.50    \\
                                & týr-the-Pruner &  \XSolidBrush &9.96 & 53.49 \\
                               & IntraSlice & \XSolidBrush & 7.72	&	\textbf{58.86} \\ 
         \bottomrule
    \end{tabular}}

    \caption{Comparison results of our method with LLM-surgeon, týr-the-Prune and DISP-LLM methods. In order to align with DISP-LLM, we only tested 5 zero-shot tasks. The PPL is tested on wikitext2 with sequence length 4096. E-T means end-to-end training.}
    \label{table2}
\end{table}

\subsubsection{Implementation details.}
 All experiments are conducted with PyTorch using the Hugging Face Transformers library on a single NVIDIA A800 GPU, with an exception for the 70B model which requires two GPUs. Following the SoBP\cite{wei2024sobp} and SliceGPT \cite{ashkboos2024slicegpt}, we use a calibration set of 128 samples from the WikiText2 \cite{merity2017wikitext2} training set, with a sequence length of 2048. We also evaluate on the C4 dataset to further validate the effectiveness of our approach (see Appendix \ref{sec:appendix_c4} for C4 results).

\subsection{Main Pruning Results}
The perplexity (PPL) and zero-shot task accuracies of compressed models using different methods are reported in Table \ref{table1}. Our IntraSlice consistently outperforms other approaches in most experiments. Compared to SliceGPT \cite{ashkboos2024slicegpt}, an inter-PCA compression method, our intra-PCA-based method demonstrates clear advantages, particularly under high sparsity.
It is worth noting that SliceGPT \cite{ashkboos2024slicegpt} introduces additional parameters at each residual connection, which undermines its acceleration efficiency. 
In comparison to the latest state-of-the-art method, SoBP\cite{wei2024sobp}, our approach exhibits stable improvements across well-trained models such as LLaMA2-7B, LLaMA2-13B, and LLaMA3-8B. Only in the LLaMA3-8B at 40\% sparsity does the zero-shot accuracy slightly decrease.
For the larger model LLaMA2-70B, our IntraSlice method substantially improves zero-shot task performance, while achieving slightly better perplexity than the baseline. 
We further tested at higher sparsity levels, as shown in Table \ref{table_high_sparse}. When the sparsity reaches 50\% and 60\%, our IntraSlice shows a more significant improvement over SoBP.

In Table \ref{table2}, we present a detailed comparison between IntraSlice, DISP-LLM, týr-the-Pruner, and LLM-Surgeon on LLaMA2-7B and LLaMA2-13B. The proposed IntraSlice achieves consistent perplexity and zero-shot accuracy improvement over týr-the-Pruner and LLM-Surgeon. When compared with DISP-LLM, our IntraSlice achieves remarkably higher zero-shot accuracy for both models at all the sparsity levels. Additionally, while DISP-LLM performs well in reducing perplexity, their performance in zero-shot tasks is poor compared with LLM-Surgeon and our IntraSlice. We attribute this to its end-to-end training paradigm and more training data (2$\times$10000 samples), which highlights the potential of overfitting associated with DISP-LLM. Additionally, under lower sparsity levels, our method outperforms DISP-LLM, even using only 128 calibration samples. At the same time, in zero-shot tasks, IntraSlice consistently outperforms DISP-LLM by a large margin at all sparsity levels, which shows the strong generalization ability of the proposed IntraSlice.


\subsection{Speedup Test}
\begin{figure}[t]
    \centering
    \includegraphics[width=1.0\columnwidth]{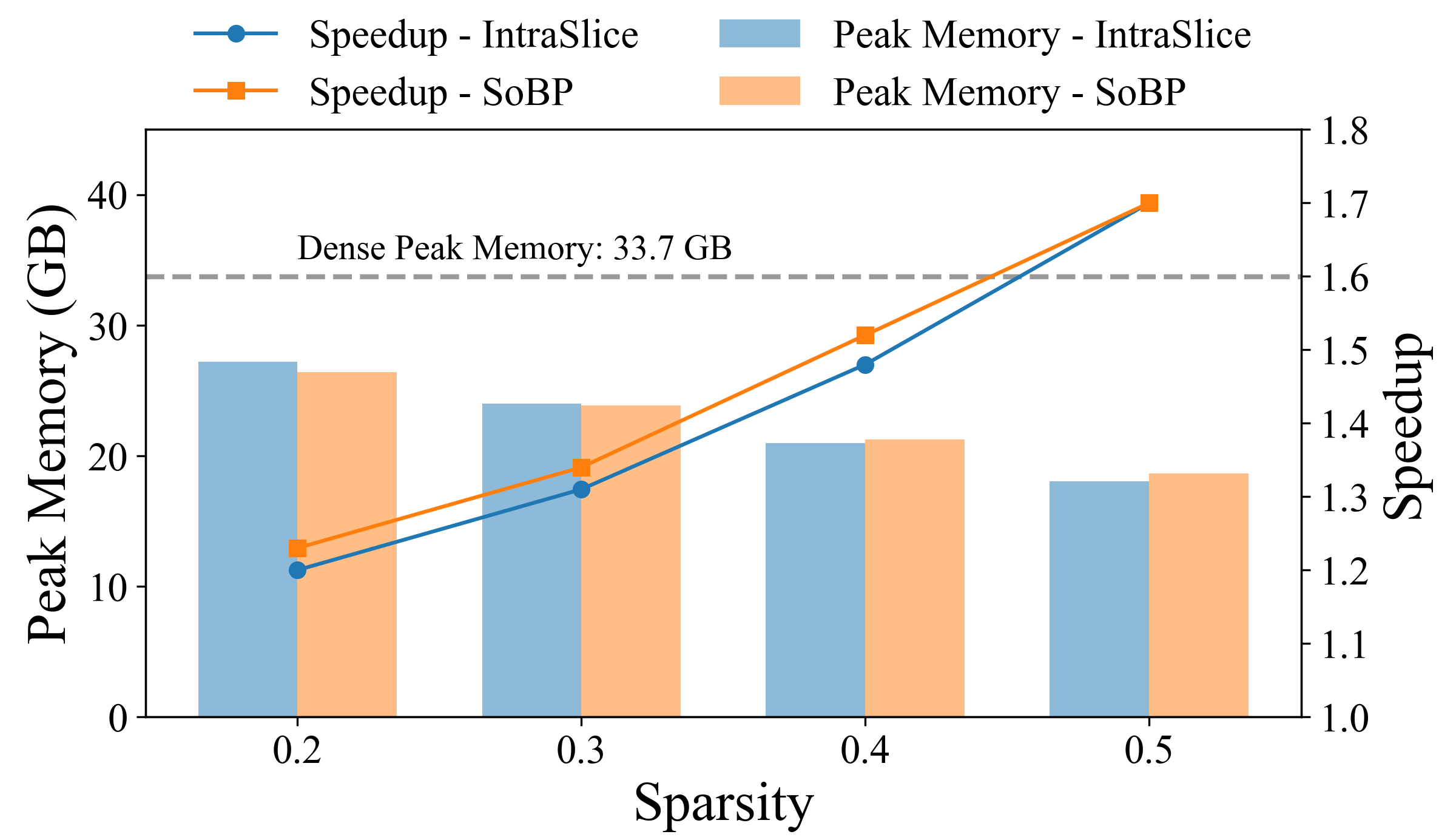} 
    \caption{Inference speedup and memory of SoBP and IntraSlice at different sparsities, on the llama2-13B model.}
    \label{fig_end}
     \vspace{-6pt}
\end{figure}
The fusion mechanism of our approach is illustrated in Figure \ref{fig3}. For models based on the transformer architecture, our method can be fully integrated into the model weights without requiring additional computations. More details of matrix fusion can be found in Appendix \ref{sec:appendix_fuse}.

The acceleration performance of LLaMA2-13B using our method is shown in Figure \ref{fig_end}. Our method is comparable to current state-of-the-art structured pruning methods in terms of acceleration and memory usage, but IntraSlice can achieve better compression performance. More details of speedup can be found in Appendix \ref{sec:appendix_speedup}.

\subsection{Ablation Study}
\begin{table}[t]
    \centering
    \small
     \setlength{\tabcolsep}{1.2mm}
    \renewcommand{\arraystretch}{0.8}
    {
    \begin{tabular}{ccc|c|c}
        \toprule    
        \multirow{2}{*}{\textbf{Ada-H}}&\multirow{2}{*}{\textbf{Pca-A}}&  \multirow{2}{*}{\textbf{IPca}} &  \textbf{20\% Sparsity} &\textbf{30\% Sparsity}  \\
            
            \cline{4-5}
           &    &  & PPL$\downarrow$   & PPL$\downarrow$   \\
         \midrule
          \XSolidBrush & \XSolidBrush & \XSolidBrush &  6.54   & 7.74   \\
           \Checkmark & \XSolidBrush & \XSolidBrush &  6.32   & 7.28    \\
        \Checkmark &\Checkmark & \XSolidBrush &  6.29 & 7.14    \\
       \Checkmark & \Checkmark & \Checkmark &  \textbf{6.27}   & \textbf{7.11}   \\

        \bottomrule
            
    \end{tabular}}
    \caption{Impact of each component on Llama2-7B.}
    \label{table4}
\end{table}
\subsubsection{Impact of components.}

We perform ablation studies to evaluate the impact of each component. Ada-H refers to the application of an adaptive PCA-based head pruning strategy. Pca-A uses a sparse PCA correction based on gradient evaluation when estimating the global pruning ratio, while IPca applies an iterative optimization process on the results of direct initialization within the FFN module. The naive baseline consists of global non-uniform pruning based on only mask gradients, un-iterated initialization of $Q_c$ and $Q_r$ in FFN, and traditional entire head pruning in MHA.
As shown in Table \ref{table4}, Ada-H significantly improves PPL performance. While Pca-A and IPca achieve relatively small improvements compared to Ada-H, they both consistently improve compression performance. Notably, Pca-A enhances sensitivity to important heads, reducing the effect of redundancy during pruning evaluation.
\subsubsection{Impact of pruning bias $\lambda_b$.}
We introduce a pruning bias $\lambda_b$ to control the pruning tendency. As shown in Figure \ref{fig5} (b), we tested the impact of the pruning bias on PPL and zero-shot accuracy. When $\lambda_b$ is 1.0, that is completely balanced, it is comparable to the best result. The $\lambda_b$ settings used in our paper can be found in Appendix \ref{sec:appendix_bias}.
\subsubsection{Impact of the number of calibrations.}
\begin{figure}[t]
    \centering
    \includegraphics[width=1.0\columnwidth]
    {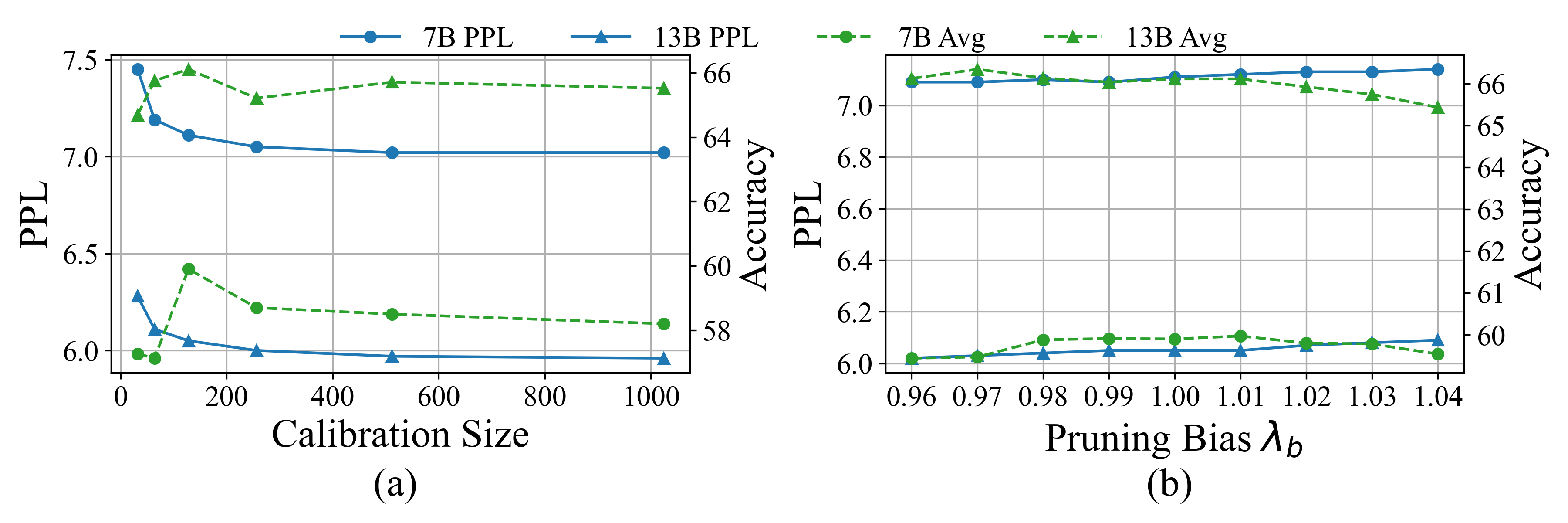} 
    \caption{The impact of the size of calibration data and pruning bias $\lambda_b$ on PPL and zero-shot accuracy on Llama2-7B and Llama2-13B with 30\% sparsity.}
    \label{fig5}
\end{figure}

As shown in Figure \ref{fig5} (a), we evaluate the effect of the number of calibration data on the results. As the number of calibration data increases, PPL decreases steadily until it starts to slow down around 1024. However, the accuracy of zero-shot tasks does not consistently improve with the increase in the amount of calibration data.

\section{Conclusion}

In this work, we propose a novel structured pruning framework for large language models that leverages block-level intra-module PCA to alleviate the performance degradation commonly seen in traditional pruning methods.
To address the integration challenges of intra-PCA within MHA and FFN components, IntraSlice introduces an adaptive, head-wise PCA-based compression strategy along with a progressive, sliced iterative PCA. This design facilitates the seamless fusion of intra-PCA transformation matrices into model weights while retaining strong compression effectiveness.
To better estimate the global non-uniform pruning ratio, we apply sparse PCA to capture post-compression activation patterns, enabling more accurate estimation of the global non-uniform pruning ratio.
Extensive experiments demonstrate that IntraSlice achieves state-of-the-art results across multiple models, delivering significant improvements in pruning performance compared to existing methods. 
We believe this PCA-driven framework is a key step toward high-performance model compression. In our future work, we will aim to develop more flexible and efficient intra-PCA pruning strategies.

\section*{Limitations}
While IntraSlice is effective, it has limitations. First, the iterative PCA introduces modest computational overhead when pruning, especially for large models, and its optimization remains suboptimal. Solving PCA under nonlinear structural constraints in transformers is still challenging. Second, for attention variants like Grouped Query Attention (GQA), query and key projections must be compressed consistently within each group to enable matrix fusion, imposing structural constraints that limit pruning flexibility. Addressing these challenges without sacrificing performance is a promising research direction.


\bibliography{custom}

\clearpage  
\appendix

\section{Rank Computation of Activation}
\label{sec:appendixA}
To evaluate the rank of activations in large language models, we adopt the widely used Energy-Based Rank Selection method with a threshold of 99\%. As illustrated in Figure 1, the activation distribution in the early layers is largely dominated by the input embeddings. Due to their uniformly distributed channels, these embeddings exhibit a high rank. In the first few transformer layers, however, outlier channels begin to emerge and dominate the activations, leading to a sharp drop in rank. As computation proceeds, the number of informative outlier channels gradually increases, resulting in a slow recovery of the rank.

To integrate the compression matrix into the model weights, inter-module PCA methods such as SliceGPT require compressing both the residual path and the module output. This leads to significant distributional shifts that can propagate to downstream layers via residual connections. In contrast, our method, IntraSlice, only modifies the output of the current module. This minimizes disruption to the overall activation distribution and has a negligible impact on subsequent layers.

\section{Algorithm}
Algorithm \ref{alg:algorithm} presents an overview of the IntraSlice framework. Algorithm \ref{alg:head_algorithm} details the adaptive head compression procedure for the MHA module, while Algorithm \ref{alg:ffn_algorithm} outlines the progressive iterative PCA compression applied to the FFN module. For clarity and concreteness, we illustrate the implementation using the LLaMA architecture as an example.
\begin{algorithm}[th]
\small
\caption{Prune Algorithm of IntraSlice}
\label{alg:algorithm}
\textbf{Input}: Sparsity $r$, Calibration $\mathcal{D}$, Model\\
\textbf{Parameter}: Pruning bias $\lambda_b$\\
\textbf{Output}: Model after pruning
\begin{algorithmic}[1] 
\STATE \textbf{Part 1: Global non-uniform prune ratio evaluation }
\STATE $G$, $X$ $\gets CrossEntropyLoss$(Model, $\mathcal{D}$)  
\STATE $Q_S \gets SparsePCA(X)$
\STATE $I \gets (GQ_s)^2$
\STATE $S^H=\{S^{H}_1,...,S^H_L\}$, $S^F=\{S^{F}_1,...,S^F_L\}$ \\$
\gets GetGlobalPruningRatio(I, \lambda_b, r)$ 
\FOR{$l = 1$ \TO $L$}
\STATE \textbf{Part 2: Adaptive head PCA-based compression }
\STATE $mha \gets l^{th}$ MHA module 
\STATE $X^H_l \gets$ inputs of  $l^{th}$ MHA module
\STATE $mha \gets$ $Algorithm$\ref{alg:head_algorithm}($S^H_l$, $mha$, $X_l^H$)
\STATE \textbf{Part 3: Progressive sliced iterative PCA }
\STATE $ffn \gets l^{th}$ FFN module 
\STATE $X_l^F \gets$ inputs of  $l^{th}$ FFN module
\STATE $ffn \gets$ $Algorithm$\ref{alg:ffn_algorithm}($S^F_l$, $ffn$, $X_l^F$)

\ENDFOR

\STATE \textbf{return} solution
\end{algorithmic}
\end{algorithm}

\begin{algorithm}[th]
\small
\caption{Adaptive Head PCA-based Compression}
\label{alg:head_algorithm}
\textbf{Input}: Sparsity $r$, Inputs $X$, MHA 
\begin{algorithmic}[1] 
\STATE $q\_proj$, $k\_proj$, $v\_proj$, $o\_proj \gets$ MHA 
\STATE \textbf{\# Compression of v and o}
\STATE $X_o \gets$ inputs of $o\_proj$ 
\STATE $V \gets$ PCA of $X_o$ 
\STATE $W_o \gets$ weights of  $o\_proj$ 
\STATE $W_v \gets$ weights of  $v\_proj$
\STATE $R \gets$  $HeadReconstructionScore$($X_o$, $W_o$) \hfill  $ \rhd$ Eq.1  
\STATE $remove\_head \gets$ $\varnothing$           
\FOR{$\_ = 1$ \TO $H$}
\STATE    $h_r \gets$  head with lowest score $R$
\STATE    $S_g \gets$ $ReconstructionScoreGain$($h_r$, $V$,$R$, $r$)\\ \hfill                $ \rhd$ Eq.3,4 
\IF{$S_g > 0$}
\STATE  $remove\_head \gets remove\_head \cup \{ h_r \}$ 
\ENDIF
\ENDFOR
\STATE $p \gets$  head dim after compression
\STATE $Q_2 \gets$ PCA($X_o$, $p$) 
\STATE $Q_2^* \gets$ $Correct$($Q_2^T$, $X_o$)   \hfill      $ \rhd$ Eq.5 
\STATE $W_o \gets$ $Fusion Matrix$($W_o$, $Q_2^*$)
\STATE $W_v \gets$ $Fusion Matrix$($Q_2$, $W_v$)
\STATE \textbf{\# Compression of q and k}
\STATE $Out_q$, $Out_k$ $\gets$ outputs of $q\_proj$ and $k\_proj$
\STATE $Q_1 \gets$ PCA($Out_q$, $Out_k$, $p$, $remove\_head$)
\STATE $W_q \gets$ weights of  $q\_proj$ 
\STATE $W_k\gets$ weights of  $k\_proj$
\STATE $W_q \gets$ $Fusion Matrix$($Q_1$, $W_q$)    
\STATE $W_k \gets$ $Fusion Matrix$($Q_1$, $W_k$)    
\STATE \textbf{return} MHA
\end{algorithmic}
\end{algorithm}

\begin{algorithm}[th]
\caption{Progressive Sliced Iterative PCA}
\small
\label{alg:ffn_algorithm}
\textbf{Input}: Sparsity $r$, Inputs $X$, FFN
\begin{algorithmic}[1] 
\STATE $up\_proj$, $gate\_proj$, $down\_proj\gets$ FFN
\STATE $W_u \gets$ weights of  $up\_proj$ 
\STATE $W_g\gets$ weights of  $gate\_proj$
\STATE $W_d\gets$ weights of $down\_proj$
\STATE $X_d \gets$ inputs of $down\_proj$
\STATE $Out_u$, $Out_g \gets$  outputs of $up\_proj$ and $gate\_proj$
\STATE \# Initialize $Q_c$, $Q_r$
\STATE $Q_c \gets SelectMatrix$($X_d$, $W_d$, $r$)
\STATE $Q_r \gets OptimizeQ_r$($Q_c$, $X_d$) \hfill  $ \rhd$ Eq.8
\IF{Need Iteration}
\FOR{$i=1$ \TO $iteration\_number$}
\STATE $Q_c \gets SliceOptimizeQ_c$($Q_c$, $Q_r$, $Out_u$, $Out_g$, $X_d$)  \hfill $\rhd$ Eq.7
\STATE $Q_r \gets OptimizeQ_r$($Q_c$, $X_d$) \hfill  $ \rhd$ Eq.8
\ENDFOR
\ENDIF
\STATE $W_u \gets$ $Fusion Matrix$($Q_c$, $W_u$)    
\STATE $W_g \gets$ $Fusion Matrix$($Q_c$, $W_g$)
\STATE $W_d \gets$ $Fusion Matrix$($W_d$, $Q_r$)    
\STATE \textbf{return} FFN
\end{algorithmic}
\end{algorithm}

\section{Analysis of MHA matrix fusion}
\label{sec:appendix_fuse}
During pruning of the MHA module, IntraSlice eliminates low-information heads based on PCA analysis and applies uniform compression to the remaining heads to achieve effective compression. Since this process alters the dimensionality of attention heads, different fusion strategies are required depending on the specific MHA architecture. Nonetheless, all variants support full integration of the compressed structure.

\begin{figure}[t]
    \centering
    \includegraphics[width=0.9\linewidth]{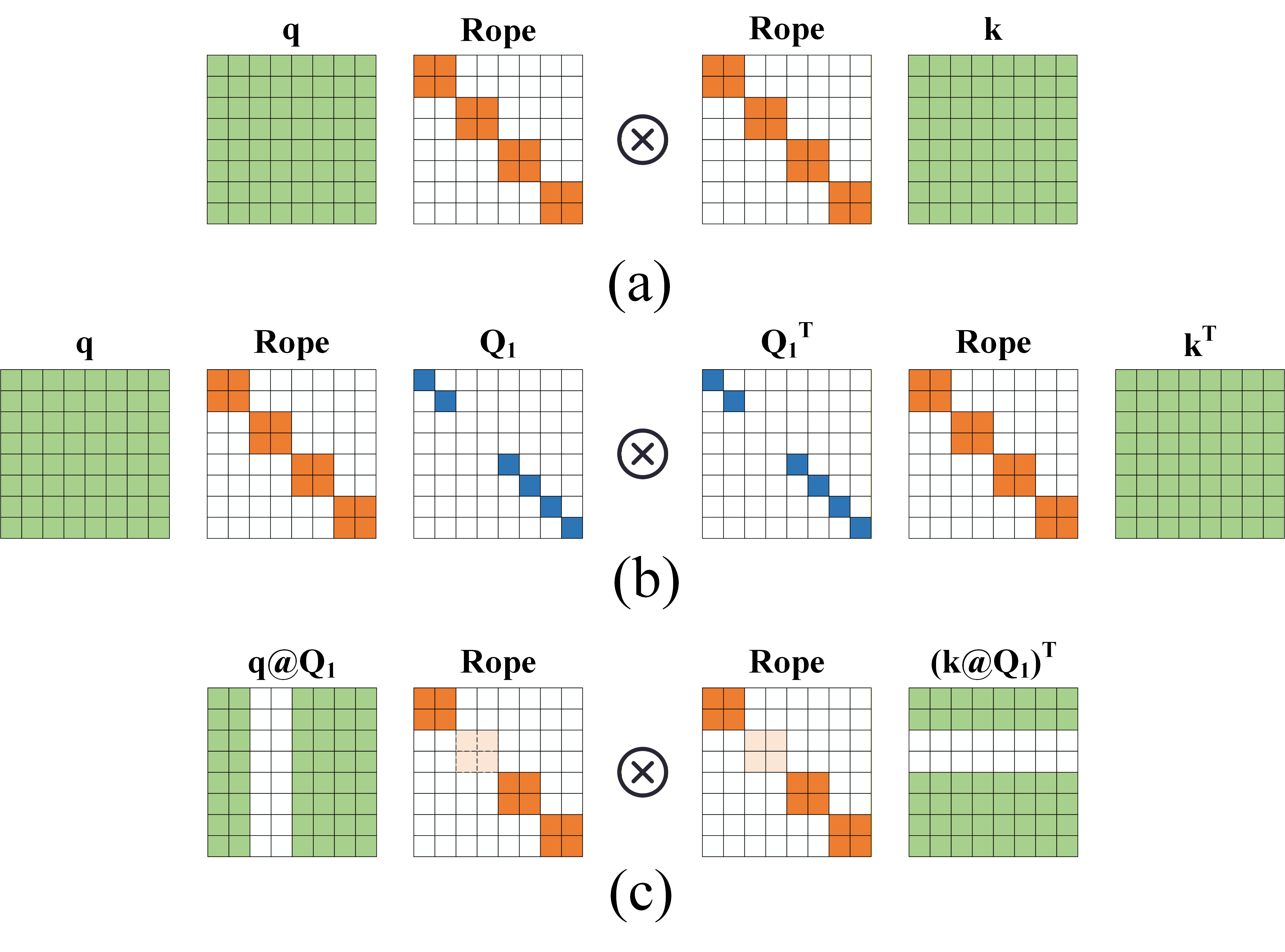}
    \caption{Schematic diagram of query and key compression with ROPE. $Q_1$ is the pairwise selection matrix. (a) Calculation of $q\times k$ with ROPE; (b) Construction of pairwise selection matrix $Q_1$; (c) Fusion $Q_1$ into query and key weights.}
    \label{fig:rope}
\end{figure}

\begin{figure}[t]
    \centering
    \includegraphics[width=0.9\linewidth]{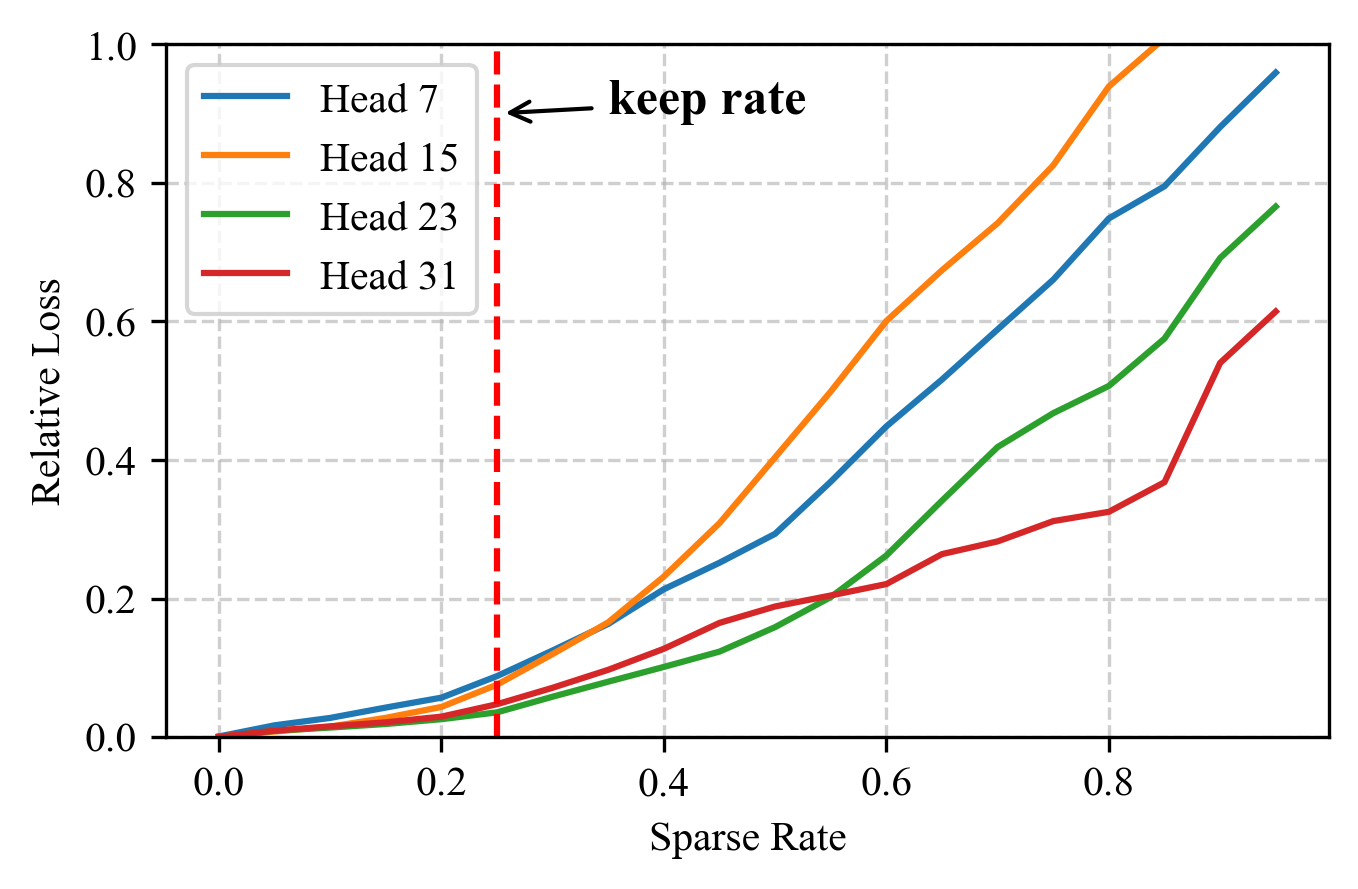}
    \caption{On Llama2-7b, 5th layer, the relative loss of attention output for different heads under different sparsity in ROPE with pairwise channel selection matrix.}
    \label{fig:rope_loss}
\end{figure}
\subsection{Query and Key Compression with ROPE}
\label{sec:appendix_rope}
RoPE applies distinct rotational position encodings to tokens at different positions. Conventional compression transformation matrices cannot pass through the RoPE operation and thus cannot be fused into the model weights. To enable matrix fusion, we construct a pairwise channel selection matrix adapted to the structure of RoPE. By leveraging its pairwise encoding properties, we assess channel importance and prune unimportant channels in pairs, as illustrated in Figure \ref{fig:rope}. We tested RoPE sparsity's effect on attention output in Llama2-7b. As shown in Figure \ref{fig:rope_loss}.  At 25\% sparsity, average output relative error is 7.43\%; at 50\%, 27.8\%. When the sparsity is within 25\% (Keep Rate), the sparsity of the rope has a small impact on the output. We set minimum dimension of “compressed head” to 96 (original 128) to preserve performance.

In models with ROPE, where positional encoding is tied to specific channels, a minimal binary mask is introduced to track pruned query and key channels, which is negligible in practice.

\begin{figure}[th]
    \centering
    \includegraphics[width=1.0\linewidth]{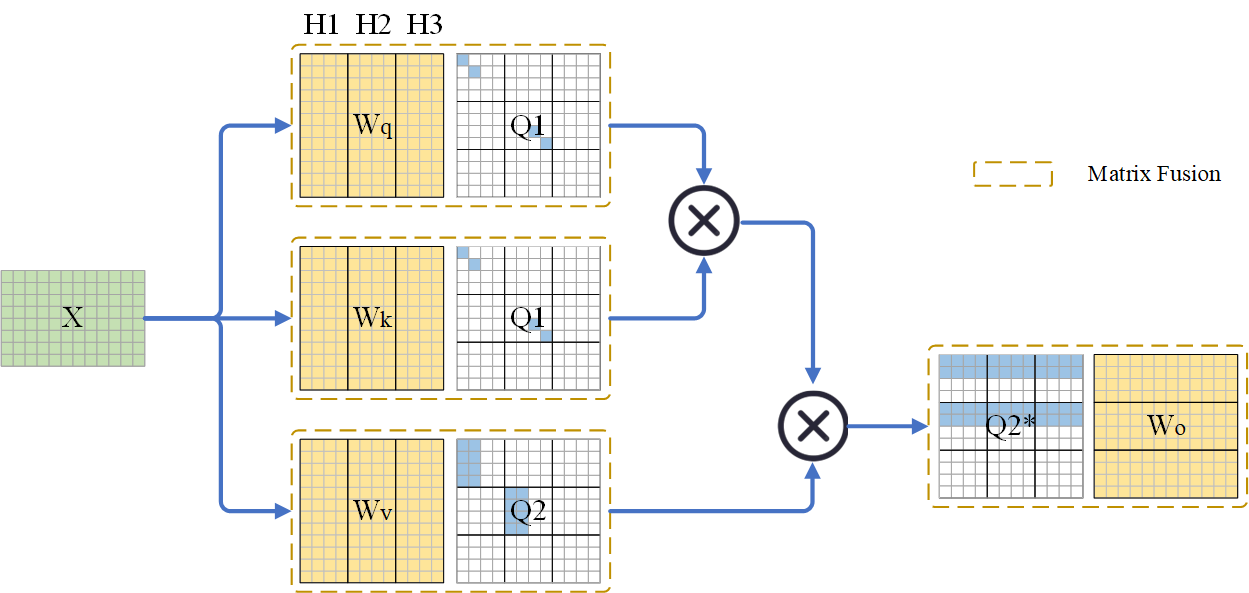}
    \caption{Schematic diagram of fusion $Q_1$, $Q_2$ and $Q_2^*$ into llama MHA weights with ROPE. }
    \label{fig:llama}
\end{figure}
\subsection{Matrix Fusion of Llama Family}
For the MHA of the ROPE llama architecture, the fusion of its transformation matrix is shown in Figure \ref{fig:llama}.

\begin{figure}[th]
    \centering
    \includegraphics[width=1.0\linewidth]{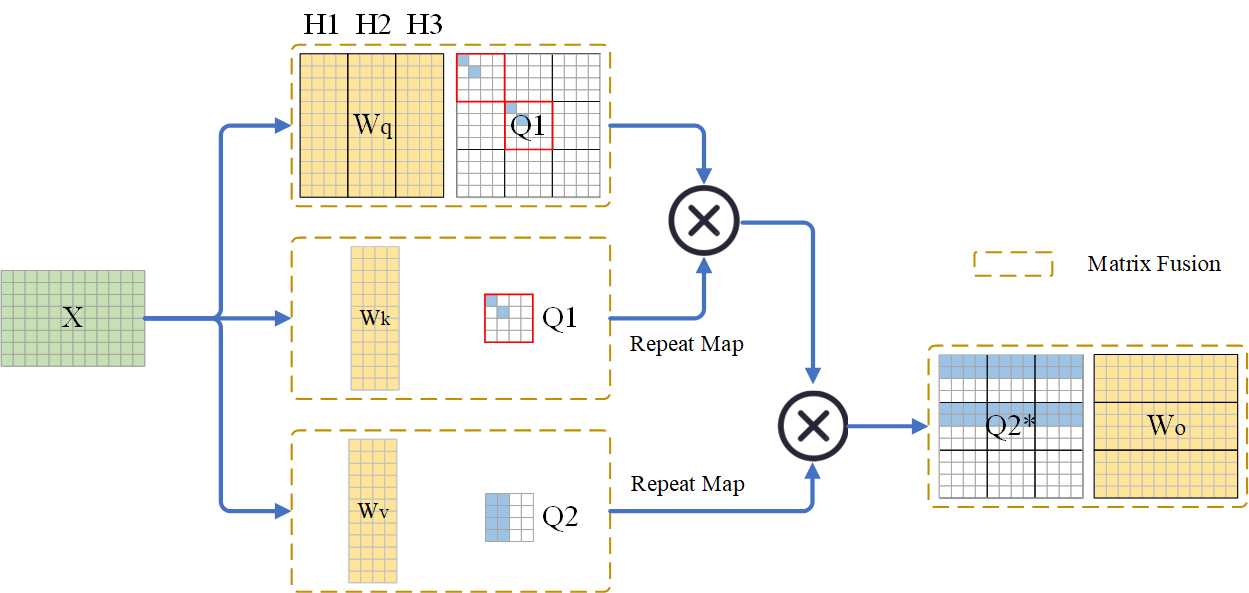}
    \caption{Schematic diagram of fusion $Q_1$, $Q_2$ and $Q_2^*$ into llama MHA weights with ROPE and GQA.}
    \label{fig:llama_group}
\end{figure}

\subsection{Matrix Fusion of Llama Family with GQA}
As shown in Figure \ref{fig:llama_group}, for MHA using GQA, the query and key $Q_1$ of a head within a group are the same, marked with a red rectangle. The construction of $Q_2^*$ is unaffected; only group-wide analysis is required when constructing $Q_2$.

\begin{table*}[!t]
\centering
\small
\setlength{\tabcolsep}{2pt}
\renewcommand{\arraystretch}{0.85}
\begin{tabular}{c|c|c|ccccccc|c}
\toprule
\multirow{2}{*}{\textbf{Sparsity}} & \multirow{2}{*}{\textbf{Model}} & \multirow{2}{*}{\textbf{PPL}} & \textbf{WinoGrande} & \textbf{PIQA} & \textbf{OBQA} & \textbf{hellaswag} & \textbf{BoolQ} & \textbf{ARC\_e} & \textbf{ARC\_c} & \multirow{2}{*}{\textbf{Avg}} \\
 \cline{4-10}
 &  &  & acc & acc-norm & acc-norm &   acc-norm & acc & acc-norm & acc-norm &  \\
\midrule
 0\% & Llama2-7b & 5.47 & 68.98 & 79.05 & 44.2 & 76.02 & 77.74 & 74.58 & 46.25 & 66.69 \\
\midrule
\multirow{2}{*}{20\%}   & SoBP & 9.36 & 64.48 & 76.22 & 37.4 & 68.99 & 70.7 & 68.56 & 41.13 & 61.07 \\
  & IntraSlice & \textbf{7.11} & \textbf{68.43} & \textbf{76.88} & \textbf{40.2} & \textbf{72.68} & \textbf{75.66} & \textbf{69.02} & \textbf{41.98} & \textbf{63.55} \\
\midrule
\multirow{2}{*}{30\%} 
  & SoBP & 12.82 & 62.35 & 73.5 & 34.6 & 65.04 & 70.52 & 61.45 & 38.57 & 58.00 \\
  & IntraSlice & \textbf{8.87} & \textbf{67.64} & \textbf{75.08} & \textbf{38.4} & \textbf{68.84} & \textbf{75.2} & \textbf{65.03} & \textbf{38.05} & \textbf{61.18} \\
\midrule
\midrule
 
 0\% & Llama2-13b & 4.88 & 72.3 & 80.52 & 45.2 & 79.38 & 80.58 & 77.53 & 49.23 & 69.25 \\
\midrule
\multirow{2}{*}{20\%}  & SoBP & 8.5 & 70.56 & 77.91 & 42.8 & 77.23 & 80.73 & 74.41 & 46.25 & 67.13 \\
  & IntraSlice & \textbf{5.86} & \textbf{72.38} & \textbf{79.49} & \textbf{44.0} & \textbf{77.49} & \textbf{81.96} & \textbf{74.92} & \textbf{47.27} & \textbf{68.21} \\
\midrule
\multirow{2}{*}{30\%} 
  & SoBP & 11.0 & 69.69 & 76.93 & 39.6 & 73.07 & 80.73 & 70.62 & 43.77 & 64.92 \\
  & IntraSlice & \textbf{6.99} & \textbf{71.11} & \textbf{77.64} & \textbf{41.8} & \textbf{75.35} & \textbf{81.31} & \textbf{71.38} & \textbf{45.31} & \textbf{66.27} \\
\bottomrule
\end{tabular}
\caption{Comparison of model compression results of different methods with the C4 calibration dataset.}
\label{table6}
\end{table*}

\section{More Ablation Study}
\subsection{Ablation Study of C4 Dataset}
\label{sec:appendix_c4}
To further verify the effectiveness of our method, we tested it on the C4 dataset and compared it with the current SOTA method SoBP. To maintain consistency in the settings, we use a calibration set of 128 samples from C4 train set with a length of 2048. PPL was tested on Wikitext2 with a length of 2048.


As shown in Table \ref{table6}, even when evaluated on the C4 dataset, IntraSlice consistently achieves better perplexity (PPL) compared to SoBP. Across both LLaMA2-7B and LLaMA2-13B, our method significantly outperforms SoBP in terms of both PPL and zero-shot accuracy. These results highlight the effectiveness and strong generalization capability of IntraSlice.


\begin{table}[h]
\centering
\small
\renewcommand{\arraystretch}{1.0}
\begin{tabular}{lccc}
\hline
\textbf{Model} & \textbf{20\%} & \textbf{30\%} & \textbf{40\%} \\
\hline
LLaMA2-7B & 1.00 & 1.00 & 1.00 \\
LLaMA2-13B & 1.00 & 0.82 & 1.00 \\
LLaMA2-70B & 0.82 & 0.91 & 0.94 \\
LLaMA3-8B & 0.98 & 0.95 & 1.00 \\
Phi-3-medium-4k & 1.00 & 1.00 & 1.00 \\
\hline
\end{tabular}
\caption{Setup of $\lambda_b$ of different model and different sparsity.}
\label{table5}
\end{table}

\subsection{Setup of Prune Bias $\lambda_b$}
\label{sec:appendix_bias}
When constructing the global non-uniform pruning ratio, we introduce a pruning bias to adjust the relative sparsity between the MHA and FFN modules. While setting this bias to 1 yields good results for most models and sparsity levels, certain models benefit from a more tailored setting to achieve optimal performance. Table \ref{table5} summarizes the bias configurations used for different models and sparsity levels in this work.

\section{Compression Time}
\begin{table}[th]
    \centering
     \setlength{\tabcolsep}{1.2mm}
    \renewcommand{\arraystretch}{0.8}
    \small
    {
    \begin{tabular}{c|c|c|c}
        \toprule
        \textbf{Method} & Llama2-7B & Llama2-13B & Llama2-70B \\
        \midrule
        SliceGPT   & 0.20h & 0.23h & 1.97h \\
        SoBP      & 0.46h & 0.82h & 8.54h \\
        IntraSlice  & 0.52h & 0.97h & 7.38h \\
        týr-the-Prune & 13.06h & 27.93h & -- \\
        DISP-LLM  & 4.82h & 35.32h & -- \\
        LLM-Surgeon  & 68.52h & 267.44h & -- \\
        \bottomrule
            
    \end{tabular}}
    \caption{Comparison of compression time (GPU hours) of different methods on various models. SliceGPT, SoBPand IntraSlice are executed on the A800, while DISP-LLM and LLM-Surgeon are executed on the A100 and H100, respectively.}
    \label{table3}
\end{table}

\begin{table*}[ht]
   \small
    \setlength{\tabcolsep}{3pt}
    \renewcommand{\arraystretch}{0.85}
    \centering
    {
    \begin{tabular}{c|c|c|ccccc|c}
         \toprule
       \multirow{2}{*}{\textbf{Sparsity}} & \multirow{2}{*}{\textbf{Method}} &\multirow{2}{*}{\textbf{PPL}} & \textbf{WinoGrande} &  \textbf{PIQA} & \textbf{HellaSwag} & \textbf{ARC-e} & \textbf{ARC-c} &  \multirow{2}{*}{\textbf{Avg}} \\
            
            \cline{4-8}
                           &    &   & acc  & acc-norm  & acc-norm  & acc-norm  & acc-norm  &   \\
         \midrule
         
           0\%        & Llama2-7B&  5.12 &   69.14    &     79.11    &  75.99    &    74.58   &   46.15      &   68.99\\
         \midrule    
         \multirow{3}{*}{30\%} & LLM-Surgeon &  7.83 &  61.09 & 73.56 & 60.72 & 63.09 & 36.69 & 59.03   \\
                               & DISP-LLM   &    6.85 &  62.27& 71.82 &63.43 &59.81 &33.19 & 58.10    \\
                               & týr-the-Pruner & 7.00  & 64.17 & 74.27 & 66.41 & 52.74 & 32.00 & 57.92 \\
                               & IntraSlice &  \textbf{6.61}   & \textbf{66.93} &  \textbf{74.59} & \textbf{68.16} &  \textbf{63.59} & \textbf{32.23 } &   \textbf{62.13}   \\
         \midrule
         \multirow{3}{*}{50\%} & LLM-Surgeon&   15.38 &  52.57 &64.36& 40.29 &44.91 &26.28 & 45.68    \\
                               & DISP-LLM   & \textbf{9.84}           &  54.54 &63.93 &46.33 &43.06 &25.85 & 46.72    \\
                               & týr-the-Pruner & 10.44 & 55.64 & 65.89 & 50.83 & 45.37 & 27.73 & 49.10\\
                               & IntraSlice &   10.11 &  \textbf{61.09} & \textbf{64.8} &  \textbf{50.21} & \textbf{48.36} &  \textbf{29.78} & \textbf{50.85}  \\
       \midrule
       \midrule

        0\% & Llama2-13B& 4.25	&80.52& 72.3&	79.38&	77.53&	49.23	&	71.79 \\
         \midrule

         \multirow{4}{*}{30\%} & LLM-Surgeon  & 6.21 &	68.67 &	76.5 & 71.52 &69.74	 &40.27	 &65.34    \\
                                & DISP-LLM  & 5.77 & 74.43 &	66.85 &	70.86 &	63.8 &	39.42	 &	63.07    \\
                                 & týr-the-Pruner & 6.05 & 69.06 & 76.93 & 72.66 & 64.48 & 40.61 & 64.75 \\
                               & IntraSlice &  \textbf{5.63} & \textbf{76.71} &	\textbf{71.98} &	\textbf{73.87} &	\textbf{73.91} &	\textbf{44.88} &		\textbf{68.27}  \\ 
                               \midrule

        \multirow{4}{*}{50\%} &  LLM-Surgeon  & 9.43 &	68.77 & 63.22 &	56.19 &	56.19 &	31.83 &		55.24    \\
                                & DISP-LLM  & \textbf{7.11} &	59.27&68.77&	58.63&	52.57&	33.28&		54.5    \\
                                 & týr-the-Pruner & 9.96 & 59.27 & 68.99 & 58.46 & 50.51 & 30.20 & 53.49 \\
                               & IntraSlice &  7.72	&\textbf{65.75} &\textbf{69.42}&	\textbf{60.44}	&\textbf{61.49}&	\textbf{37.2}&		\textbf{58.86  } \\ 
         \bottomrule
    \end{tabular}}

    \caption{Comparison results of our method with LLM-surgeon, týr-the-Prune and DISP-LLM methods. In order to align with DISP-LLM, we only tested 5 zero-shot tasks. The test dataset of PPL is wikitext2. And the length of wikitext2 for testing is 4096.  }
    \label{table7}
\end{table*}

As shown in Table \ref{table3}, IntraSlice achieves comparable compression times compared to recent state-of-the-art methods. The primary computational bottleneck lies in processing the intermediate dimensionality of the MLP layers. Additionally, block-wise PCA for global non-uniform pruning also contributes significantly to the overall time cost, which only needs to be computed once for all sparsity levels.
In practice, all models except LLaMA2-70B can be pruned on a single 80 GB A800 GPU. In comparison, DISP-LLM and LLM-Surgeon require 4 and 8 such GPUs for pruning LLaMA2-7B and 13B, respectively.

\section{Speedup Test}
\label{sec:appendix_speedup}
\begin{figure}[ht]
    \centering
    \includegraphics[width=1.0\columnwidth]{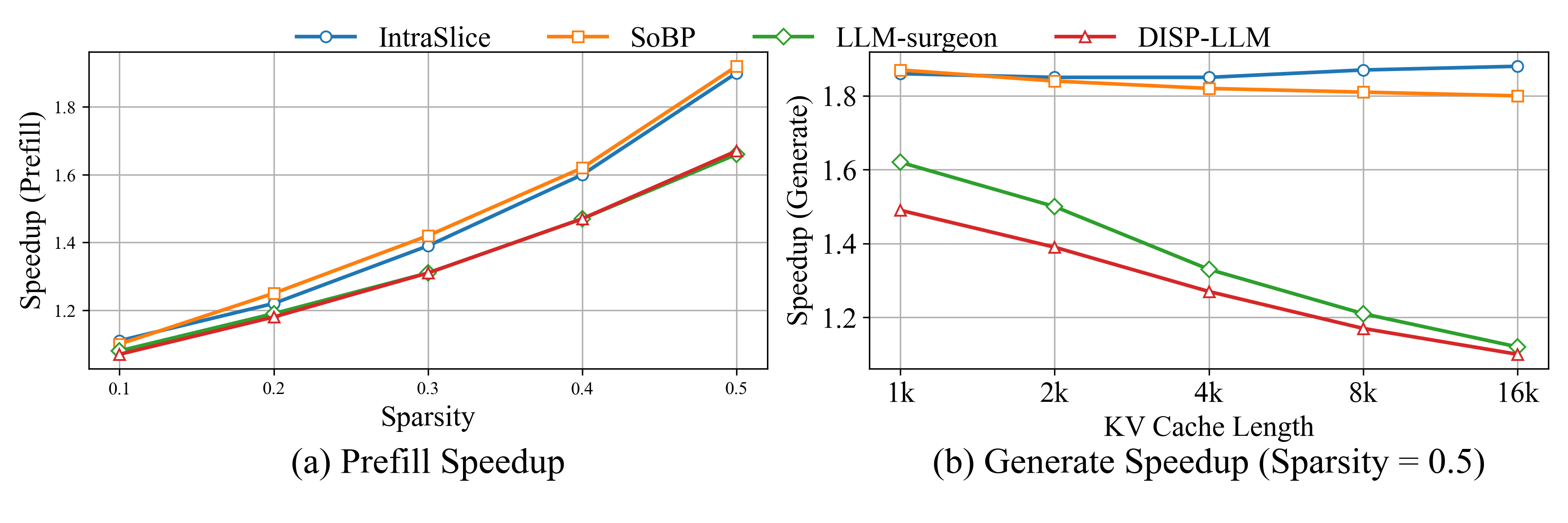} 
    \caption{Speedup of different compression methods at different sparsities, on the llama2-13B model.}
    \label{fig4}
\end{figure}
The acceleration performance of LLaMA2-13B using our method is shown in Figure \ref{fig4}. To accurately measure the speedup of different methods, we tested a single decoder layer to remove the effects of embedding and random sampling. Sparsity is the average of all layers. For all speed experiments, the prefill speed is evaluated with a 4096-length inputs. The generation speed is evaluated with a 4096-length KV cache. Compared with SoBP\cite{wei2024sobp}, our method achieves comparable or superior acceleration, while also maintaining better model performance. Compared with LLM-Surgeon \cite{llmsurgeon2024the} and DIPS-LLM \cite{gao2024disp}, IntraSlice has a significant speed advantage, especially in the case of long sequences. This is because the pruning strategies of LLM-Surgeon and DIPS-LLM cannot reduce the computational complexity of attention in the MHA.

\begin{table*}[!t]
\centering
\small
\setlength{\tabcolsep}{3pt}
\begin{tabular}{c|c|c|ccccccc|c}
\toprule
\multirow{2}{*}{\textbf{Sparsity}} & \multirow{2}{*}{\textbf{Model}} & \multirow{2}{*}{\textbf{PPL}} & \textbf{WinoGrande} & \textbf{PIQA} & \textbf{OBQA} & \textbf{hellaswag} & \textbf{BoolQ} & \textbf{ARC\_e} & \textbf{ARC\_c} & \multirow{2}{*}{\textbf{Avg}} \\
 \cline{4-10}
 &  &  & acc & acc-norm & acc-norm &   acc-norm & acc & acc-norm & acc-norm &  \\
\midrule
0\%  & \textbf{Llama2-7b} & 5.47 & 68.98 & 79.05 & 44.2 & 76.02 & 77.74 & 74.58 & 46.25 & 66.69 \\
\midrule
\multirow{5}{*}{20\%}
            & SliceGPT      & 6.86    & 64.56      & 69.21    & 40.4       & 59.02     & 48.07 & 55.05    & 35.32    & 53.09 \\ 
		     & Wanda           & 7.38    & 64.56      & 75.84    & 42         & 70.75     & 68.07 & 66.96    & 40.53    & 61.25 \\ 
		     & SVD-LLM       & 8.52    & 60.69      & 66.65    & 36.4       & 54.23     & 49.17 & 50.04    & 30.03    & 49.6  \\ 
        & SoBP        & 6.51          & \textbf{69.14} & 76.66 & 40.40 & 65.24  & \textbf{75.66} & 69.28 & 41.89 & 63.54 \\
     & IntraSlice  & \textbf{6.27} & 68.43          & \textbf{77.26} & \textbf{42.00}   & \textbf{72.27} & 73.12 & \textbf{69.70} & \textbf{43.34} & \textbf{63.73} \\
     \midrule
\multirow{5}{*}{30\%}
         & SliceGPT                 & 8.62    & 62.59      & 64.69    & 31.8       & 49.12     & 38.9  & 50.17    & 31.14    & 46.92 \\ 
		   & Wanda                    & 9.17    & 57.85      & 72.09    & 39.2       & 62.67     & 63.06 & 60.94    & 38.14    & 56.28 \\ 
		  & SVD-LLM                  & 10.95   & 58.56      & 61.59    & 34         & 45.01     & 47.13 & 42.59    & 27.13    & 45.14 \\ 
        & SoBP        & 7.59 & 66.38 & 73.45 & 38.6 & 67.36 & \textbf{70.92} & 60.48 & 38.05 & 59.32 \\
     & IntraSlice  & \textbf{7.11} & \textbf{66.69} & \textbf{74.16} & \textbf{39.6} & \textbf{67.77} & 70.28 & \textbf{66.29} & \textbf{38.65} & \textbf{60.49} \\
     \midrule
\multirow{5}{*}{40\%}
        & SliceGPT                 & 12.79   & 57.77      & 57.89    & 28.8       & 39.48     & 37.83 & 44.02    & 27.22    & 41.86 \\ 
		& Wanda                    & 14.33   & 49.96      & 62.89    & 26         & 32.74     & 57.83 & 45.29    & 26.62    & 43.05 \\ 
		 & SVD-LLM                  & 16.58   & 55.72      & 56.2     & 29.8       & 35.55     & 38.99 & 34.81    & 24.49    & 39.37 \\
        & SoBP        & 9.32 & \textbf{65.51} & 69.26 & 36.40 & \textbf{61.72} & 56.94 & 55.01 & 35.07 & 54.27 \\
     & IntraSlice  & \textbf{8.39} & 64.09 & \textbf{69.42} & \textbf{38.00}   & 60.86 & \textbf{69.33} & \textbf{57.83} & \textbf{35.15} & \textbf{56.68} \\
\midrule
\midrule
0\% & \textbf{Llama2-13b} & 4.88 & 72.3  & 80.52 & 45.2 & 79.38 & 80.58 & 77.53 & 49.23 & 69.25 \\
\midrule
\multirow{5}{*}{20\%}
     & SliceGPT                 & 6.04    & 68.35      & 71.33    & 41         & 62.79     & 44.8  & 67.3     & 41.81    & 56.76 \\ 
		 & Wanda                    & 6.66    & 69.46      & 75.68    & 42         & 64.63     & 66.33 & 67.17    & 41.72    & 61    \\ 
		 & SVD-LLM                  & 6.78    & 66.54      & 71.6     & 41.8       & 59.82     & 73.88 & 60.4     & 35.92    & 58.56 \\ 
    & SoBP        & 5.61 & \textbf{72.06} & 78.45 & 43.8 & \textbf{77.21} & 78.17 & \textbf{77.15} & 47.53 & 67.77 \\
     & IntraSlice  & \textbf{5.48} & 71.74 & \textbf{78.45} & \textbf{44.4} & 76.84 & \textbf{80.92} & 75.72 & \textbf{47.53} & \textbf{67.94} \\
     \midrule
\multirow{5}{*}{30\%} 
    & SliceGPT                 & 7.44    & 65.35      & 65.51    & 39         & 52.33     & 38.9  & 53.16    & 36.77    & 50.15 \\
		 & Wanda                    & 10.14   & 57.22      & 53.32    & 35.6       & 47.56     & 62.48 & 31.44    & 25.43    & 44.72 \\ 
		  & SVD-LLM                  & 8.21    & 64.48      & 65.89    & 36.6       & 50.29     & 66.79 & 51.56    & 29.52    & 52.16 \\ 
    & SoBP        & 6.21 & 71.35 & 76.88 & 41.8 & 74.48 & 78.23 & 74.71 & \textbf{47.27} & 66.39 \\
     & IntraSlice  & \textbf{5.96} & \textbf{71.74} & \textbf{76.88} & \textbf{45.0} & \textbf{75.04} & \textbf{79.72} & \textbf{75.21} & 46.33 & \textbf{67.13} \\
     \midrule
\multirow{5}{*}{40\%} 
    & SliceGPT                 & 10.61   & 61.25      & 59.25    & 35.8       & 42.57     & 37.83 & 44.28    & 29.78    & 44.39 \\ 
		 & Wanda                    & 21.34   & 51.38      & 56.8     & 27.8       & 34.12     & 62.08 & 32.15    & 25.09    & 41.35 \\ 
		  & SVD-LLM                  & 11.26   & 60.06      & 59.3     & 33         & 40.9      & 52.26 & 40.78    & 25.94    & 44.61 \\ 
    & SoBP        & 7.32 & 67.88 & 67.14 & 38.0 & 68.49 & 74.62 & 53.58 & 34.73 & 57.78 \\
     & IntraSlice  & \textbf{6.92} & \textbf{68.27} & \textbf{72.69} & \textbf{40.6} & \textbf{69.01} & \textbf{78.17} & \textbf{64.94} & \textbf{41.98} & \textbf{62.24} \\
\midrule
\midrule
0\% & \textbf{Llama2-70b} & 3.32 & 77.98 & 82.75 & 48.8 & 83.81 & 83.70 & 80.98 & 57.25 & 73.61 \\
\midrule
\multirow{3}{*}{20\%} 
    & Wanda                    & 4.1     & 75.77      & 80.25    & 46.8       & 81.28     & 81.68 & 77.06    & 53.16    & 70.86 \\ 
    & SoBP        & 3.91 & 76.80 & 81.50 & \textbf{49.4} & 82.97 & 68.81 & \textbf{80.81} & \textbf{58.02} & 71.19 \\
     & IntraSlice  & \textbf{3.85} & \textbf{76.95} & \textbf{81.88} & 48.0 & \textbf{83.71} & \textbf{84.16} & 79.80 & 56.14 & \textbf{72.95} \\
     \midrule
\multirow{3}{*}{30\%} 
     & Wanda                    & 4.77    & 75.22      & 79.27    & 45.4       & 79.1      & 81.5  & 75.84    & 52.99    & 69.9  \\ 
     & SoBP        & 4.41 & 76.48 & 80.36 & 47.6 & 81.57 & 67.98 & 77.95 & 53.84 & 69.40 \\
     & IntraSlice  & \textbf{4.34} & \textbf{76.87} & \textbf{80.63} & \textbf{47.8} & \textbf{82.33} & \textbf{85.02} & \textbf{79.04} & \textbf{54.18} & \textbf{72.27} \\
     \midrule
\multirow{3}{*}{40\%} 
         & Wanda                    & 4.77    & 75.22      & 79.27    & 45.4       & 79.1      & 81.5  & 75.84    & 52.99    & 69.9  \\
         & SoBP        & 4.99 & 76.16 & 78.89 & \textbf{47.6} & 79.49 & 71.53 & \textbf{75.34} & \textbf{50.60} & 68.51 \\
     & IntraSlice  & \textbf{4.91} & \textbf{76.40} & \textbf{79.27} & 46.0 & \textbf{79.49} & \textbf{84.89} & 73.95 & 49.91 & \textbf{69.99} \\
    \bottomrule
\end{tabular}
\caption{Detail comparison of model compression results on Llama2-7b, Llama2-13b and Llama2-70b. PPL is the result on wikitext2.}
\label{table8}
\end{table*}

\begin{table*}[!t]
\centering
\small
\setlength{\tabcolsep}{3pt}
\begin{tabular}{c|c|c|ccccccc|c}
\toprule
\multirow{2}{*}{\textbf{Sparsity}} & \multirow{2}{*}{\textbf{Model}} & \multirow{2}{*}{\textbf{PPL}} & \textbf{WinoGrande} & \textbf{PIQA} & \textbf{OBQA} & \textbf{hellaswag} & \textbf{BoolQ} & \textbf{ARC\_e} & \textbf{ARC\_c} & \multirow{2}{*}{\textbf{Avg}} \\
 \cline{4-10}
 &  &  & acc & acc-norm & acc-norm &   acc-norm & acc & acc-norm & acc-norm &  \\
\midrule
0\% & \textbf{Llama3-8b} & 6.13 & 72.77 & 80.74 & 45.0 & 79.13 & 81.50 & 77.74 & 53.58 & 70.07 \\
\midrule

\multirow{5}{*}{20\%} 
    & SliceGPT                 & 10.93   & 62.9       & 63.17    & 34.4       & 52.29     & 38.13 & 52.95    & 32.85    & 48.1  \\ 
     & Wanda                    & 122.41  & 49.49      & 53.54    & 25.4       & 30.77     & 50.7  & 29.67    & 23.12    & 37.53 \\ 
     & SVD-LLM                  & 47      & 53.51      & 62.79    & 30.8       & 40.93     & 58.47 & 45.2     & 26.28    & 45.43 \\ 
    & SoBP        & 8.74 & \textbf{71.82} & 75.73 & \textbf{41.6} & 71.00 & 68.07 & \textbf{73.02} & 43.69 & 63.56 \\
     & IntraSlice  & \textbf{8.27} & 70.48 & \textbf{76.39} & 41.4 & \textbf{71.30} & \textbf{78.96} & 72.64 & \textbf{45.31} & \textbf{65.21} \\
     \midrule
\multirow{5}{*}{30\%} 
       & SliceGPT                 & 17.02   & 57.7       & 56.26    & 31         & 39.67     & 37.83 & 41.08    & 26.28    & 41.4  \\ 
		   & Wanda                    & 271.71  & 48.38      & 52.61    & 25.8       & 28.46     & 50.15 & 29.25    & 20.82    & 36.5  \\
		 & SVD-LLM                  & 101.56  & 51.85      & 57.78    & 27.2       & 32.94     & 58.69 & 34.97    & 22.18    & 40.8  \\ 
      & SoBP        & 10.32& \textbf{70.09} & 72.63 & \textbf{39.8} & 64.19 & 65.17 & 61.87 & 36.43 & 58.60 \\
     & IntraSlice  & \textbf{10.25}& 67.32 & \textbf{73.39} & 39.0 & \textbf{64.20} & \textbf{70.86} & \textbf{67.63} & \textbf{42.15} & \textbf{60.65} \\
     \midrule
\multirow{5}{*}{40\%} 
    & SliceGPT                 & 30.8    & 52.96      & 53.48    & 27.2       & 32.41     & 37.83 & 35.35    & 22.53    & 37.39 \\ 
	& Wanda                    & 4258.41 & 50.51      & 50.49    & 24.2       & 26.61     & 41.41 & 26.14    & 23.38    & 34.68 \\ 
	& SVD-LLM                  & 207.99  & 50.51      & 54.52    & 26         & 29.71     & 43.58 & 31.1     & 21.59    & 36.71 \\ 
     & SoBP        & 12.48& \textbf{65.90} & 68.12 & \textbf{35.4} & \textbf{54.47} & \textbf{62.26} & 53.16 & 29.69 & \textbf{52.71} \\
     & IntraSlice  & \textbf{12.28}& 61.88 & \textbf{68.82} & 35.00 & 51.89 & 51.65 & \textbf{56.10} & \textbf{32.34} & 51.10 \\
\midrule
\midrule
0\% & \textbf{Phi-3-medium-4k} & 4.29 & 76.56 & 81.61 & 50.8 & 82.73 & 88.53 & 81.31 & 61.69 & 74.75 \\
\midrule
\multirow{5}{*}{20\%} 
    & SliceGPT                 & 6.45    & 74.03      & 73.5     & 44.4       & 67.41     & 47.16 & 75.59    & 52.05    & 62.02 \\ 
    & Wanda                    & 6.82    & 70.32      & 77.69    & 46         & 77.48     & 84.34 & 77.82    & 52.99    & 69.52 \\ 
    & SVD-LLM                  & 7.16    & 74.11      & 76.93    & 44.4       & 70.71     & 82.63 & 75.97    & 51.37    & 68.02 \\ 
    & SoBP        & 6.27 & 73.48 & 80.09 & \textbf{47.2} & 76.96 & \textbf{88.04} & 82.24 & 58.02 & 72.29 \\
     & IntraSlice  & \textbf{5.80} & \textbf{73.72} & \textbf{80.14} & 46.8 & \textbf{79.13} & 87.77 & \textbf{84.34} & \textbf{59.56} & \textbf{73.06} \\
     \midrule
\multirow{5}{*}{30\%} 
        & SliceGPT                 & 7.66    & 68.35      & 66.81    & 38.6       & 56.31     & 56.91 & 62.79    & 41.98    & 55.97 \\ 
		 & Wanda                    & 10      & 61.88      & 74.37    & 41.2       & 70.39     & 69.69 & 71       & 48.29    & 62.41 \\ 
	& SVD-LLM                  & 8.22    & 71.67      & 72.09    & 38.8       & 61.01     & 79.36 & 65.7     & 42.75    & 61.62 \\ 
      & SoBP        & 7.05 & 71.43 & 75.19 & 43.8 & 72.75 & 87.31 & 73.15 & 48.98 & 67.52 \\
     & IntraSlice  & \textbf{6.71} & \textbf{72.69} & \textbf{77.15} & \textbf{45.8} & \textbf{74.07} & \textbf{86.12} & \textbf{79.12} & \textbf{53.16} & \textbf{69.73} \\
     \midrule
\multirow{5}{*}{40\%} 
    & SliceGPT                 & 10.01   & 63.22      & 61.59    & 34.8       & 44.72     & 37.89 & 43.9     & 30.03    & 45.16 \\
		 & Wanda                    & 20.68   & 61.17      & 71.22    & 40         & 59.01     & 41.41 & 61.07    & 40.78    & 53.52 \\ 
	& SVD-LLM                  & 10.76   & 64.25      & 62.35    & 37.2       & 48.82     & 67.55 & 53.87    & 33.53    & 52.51 \\
     & SoBP        & 8.02 & \textbf{69.77} & 69.42 & \textbf{42.4} & 65.68 & \textbf{86.09} & 54.76 & 39.93 & 61.15 \\
     & IntraSlice  & \textbf{7.90} & 67.88 & \textbf{75.08} & 41.4 & \textbf{66.38} & 84.07 & \textbf{72.73} & \textbf{46.76} & \textbf{64.90} \\
    \bottomrule
\end{tabular}
\caption{Detail comparison of model compression results on Llama3-8b and Phi-3-medium-4k. PPL is the result on wikitext2.}
\label{table9}
\end{table*}

\section{Detailed Results}
\label{sec:appendix_detail}
As shown in Table~\ref{table7}, our method demonstrates a clear advantage over DISP-LLM, týr-the-Pruner and LLM-Surgeon across all five zero-shot tasks. On average, the accuracy surpasses that of DISP-LLM by 8\%. We attribute this improvement partly to DISP-LLM’s end-to-end training on Wikitext2, which may lead to a certain degree of overfitting.

Detailed comparisons of model compression results are presented in Table~\ref{table8} and Table~\ref{table9}. Our method has significant advantages over Wanda, SliceGPT and SVD-LLM. For smaller models such as LLaMA2-7B and LLaMA3-8B, IntraSlice generally outperforms SoBP, although performance varies somewhat across individual tasks. However, for larger models—including LLaMA2-13B, LLaMA2-70B, and Phi-3-Medium (14B)—IntraSlice consistently maintains a significant advantage. We believe this is because larger models contain more redundant information and have greater parameter capacity, making adaptive strategies like head pruning more effective.

\end{document}